%% file: main.tex
\documentclass[runningheads]{llncs}
\usepackage{eccv}
\usepackage{eccvabbrv}
\usepackage[T1]{fontenc}
\usepackage[utf8]{inputenc}
\usepackage{pmboxdraw}
\usepackage{graphicx}
\usepackage{booktabs}
\usepackage[accsupp]{axessibility}  %

\usepackage[colorlinks,citecolor=eccvblue]{hyperref}
\usepackage{hyperref}

\usepackage{orcidlink}

\usepackage{graphicx}

\usepackage{tikz}
\usepackage{comment}
\usepackage{amsmath,amssymb} %

\usepackage{color, colortbl}
\usepackage[numbers,sort&compress]{natbib}

\usepackage[export]{adjustbox} %
\usepackage{booktabs} %

\usepackage{amssymb}
\usepackage{mathtools}
\usepackage{stmaryrd}
\usepackage{tabularx}
\usepackage{makecell}

\usepackage{url}            %
\usepackage{booktabs}       %
\usepackage{amsfonts}       %
\usepackage{nicefrac}       %
\usepackage{microtype}      %
\usepackage{multirow}       %

\usepackage{enumitem}%
\usepackage{wrapfig}
\usepackage{comment}
\usepackage{amsmath,amssymb} %
\usepackage{algorithm}
\usepackage{algorithmic}
\usepackage{mathtools}%
\usepackage{arydshln}
\usepackage{symbols}
\usepackage{pifont}

\input{math_commands}

\newcommand*{\ShowNotes}{} %
\input{macros.tex}

\usepackage[multiple]{footmisc}

\newcommand{\myparagraph}[1]{\vspace*{0pt}{\bf #1}}

\definecolor{mybrown}{rgb}{0.87058824, 0.56078431, 0.01960784}
\definecolor{myblue}{rgb}{0.3372549 , 0.70588235, 0.91372549}
\definecolor{mypurple}{rgb}{0.8, 0.47058824, 0.7372549 }
\definecolor{myorange}{rgb}{0.835, 0.368, 0}
\definecolor{mygreen}{rgb}{0.00784314, 0.61960784, 0.45098039}
\definecolor{mygt}{rgb}{0.0078125 , 0.57421875, 0.40625}
\definecolor{mysp}{rgb}{0.84765625, 0.515625  , 0.0234375}
\definecolor{mycitecolor}{rgb}{0,0.08,0.45}
\definecolor{mygr}{rgb}{0.9607,0.9607,0.9607}
\definecolor{myoo}{rgb}{0.992,0.9176,0.9019}

\usepackage{listings}
\usepackage{authblk}

\begin{document}

\title{Learning to Obstruct Few-Shot Image Classification over Restricted Classes}

\author{Amber Yijia Zheng\thanks{indicates equal contribution} \and Chiao-An Yang\textsuperscript{$\ast$} \and
Raymond A. Yeh}
\authorrunning{A.Y.~Zheng et al.}
\institute{Department of Computer Science, Purdue University \\
\email{\tt\small \{zheng709,yang2300,rayyeh\}@purdue.edu}
}

\maketitle

\input{src/abs}

\input{src/intro}

\input{src/rel}

\input{src/prelim}

\input{src/app}

\input{src/exp}

\input{src/conc}

\section*{Acknowledgements}

This project is supported in part by an NSF Award \#2420724. 

\bibliographystyle{splncs04nat}
\bibliography{ref}

\clearpage
\input{src/supp}

\end{document}

%% file: math_commands.tex
\usepackage{amsmath,amsfonts,bm}

\def\1{\bm{1}}

\def\sp{space}

\def\vk{{\bm{k}}}

\def\vp{{\bm{p}}}

\def\vr{{\bm{r}}}

\def\vv{{\bm{v}}}

\def\vx{{\bm{x}}}
\def\vy{{\bm{y}}}

\def\mM{{\bm{M}}}

\DeclareMathAlphabet{\mathsfit}{\encodingdefault}{\sfdefault}{m}{sl}
\SetMathAlphabet{\mathsfit}{bold}{\encodingdefault}{\sfdefault}{bx}{n}
\newcommand{\tens}[1]{\bm{\mathsfit{#1}}}
\def\tA{{\tens{A}}}

\def\tF{{\tens{F}}}

\def\gA{{\mathcal{A}}}
\def\gB{{\mathcal{B}}}

\def\gD{{\mathcal{D}}}

\def\gL{{\mathcal{L}}}

\def\gP{{\mathcal{P}}}
\def\gQ{{\mathcal{Q}}}
\def\gR{{\mathcal{R}}}
\def\gS{{\mathcal{S}}}
\def\gT{{\mathcal{T}}}

\def\gY{{\mathcal{Y}}}

\def\sN{{\mathbb{N}}}

\def\sR{{\mathbb{R}}}

\DeclareMathOperator*{\argmin}{arg\,min}

%% file: macros.tex
\usepackage{color}
\usepackage{soul}
\usepackage{multirow}
\usepackage{xcolor}
\usepackage{wrapfig}

\definecolor{darkred}{rgb}{0.7,0.1,0.1}
\definecolor{darkgreen}{rgb}{0.1,0.7,0.1}
\definecolor{cyan}{rgb}{0.7,0.0,0.7}
\definecolor{dblue}{rgb}{0.2,0.2,0.8}
\definecolor{maroon}{rgb}{0.76,.13,.28}
\definecolor{burntorange}{rgb}{0.81,.33,0}
\definecolor{tealblue}{rgb}{0.212,0.459, 0.533}
\definecolor{myyellow}{rgb}{0.8627451 , 0.75294118, 0.20784314]}

\definecolor{mypink}{rgb}{0.93359375, 0.62109375, 0.83984375}

\definecolor{pp}{rgb}{0.43921569, 0.18823529, 0.62745098}
\definecolor{rr}{rgb}{0.5254902 , 0.00784314, 0.12941176}
\definecolor{bb}{rgb}{0.09019608, 0.23529412, 0.37647059}
\definecolor{yy}{rgb}{0.49803922, 0.3372549 , 0.0}
\definecolor{gg}{rgb}{0.02352941, 0.3372549 , 0.17647059}

\ifdefined\ShowNotes
  \newcommand{\colornote}[3]{{\color{#1}\bf{#2: #3}\normalfont}}
\else
  \newcommand{\colornote}[3]{}
\fi

\definecolor{lightred}{rgb}{0.9,0.4,0.4}

\definecolor{darkpink}{rgb}{0.98,0.81,0.89}

\newcommand{\bsla}{\textit{Only$\gR$}\xspace}
\newcommand{\bslb}{\textit{No$F$}\xspace}

\newcommand{\method}{LTO\xspace}
\definecolor{lto_red}{rgb}{99, 56, 55}

\definecolor{OursColor}{rgb}{1.0,0.88,0.88}

\definecolor{wolto}{rgb}{.75,0.75,0.75}

\newcommand{\centercell}[1]{\multicolumn{1}{c}{#1}}

\usepackage{xcolor}
\usepackage{soul}

\newcommand{\hlc}[2][yellow]{{%
    \colorlet{foo}{#1}%
    \sethlcolor{foo}\hl{#2}}%
}

\newlength\savewidth

\definecolor{turquoise}{cmyk}{0.65,0,0.1,0.1}
\definecolor{purple}{rgb}{0.65,0,0.65}
\definecolor{darkgreen}{rgb}{0.0, 0.5, 0.0}
\definecolor{darkred}{rgb}{0.5, 0.0, 0.0}
\definecolor{darkblue}{rgb}{0.0, 0.0, 0.5}
\definecolor{blue}{rgb}{0.0, 0.0, 1.0}
\definecolor{orange}{rgb}{1.0,0.5,0.0}

\newcommand{\hide}[1]{{}}

\renewcommand{\vec}[1]{\mathbf{#1}}

\makeatletter
\renewcommand{\paragraph}{%
  \@startsection{paragraph}{4}%
  {\z@}{0.3ex \@plus 1ex \@minus .1ex}{-1em}%
  {\normalfont\normalsize\bfseries}%
}
\makeatother

\newif\ifproofread

%% file: src/abs.tex
\begin{abstract}
Advancements in open-source pre-trained backbones make it relatively easy to fine-tune a model for new tasks. However, this lowered entry barrier poses potential risks, e.g., bad actors developing models for harmful applications. A question arises: \textit{Is possible to develop a pre-trained model that is difficult to fine-tune for certain downstream tasks?} To begin studying this, we focus on few-shot classification (FSC). Specifically, we investigate methods to make FSC more challenging for a set of restricted classes while maintaining the performance of other classes. We propose to meta-learn over the pre-trained backbone in a manner that renders it a ``poor initialization''. Our proposed Learning to Obstruct (LTO) algorithm successfully obstructs four FSC methods across three datasets, including ImageNet and CIFAR100 for image classification, as well as CelebA for attribute classification. 
\end{abstract}

%% file: src/intro.tex
\section{Introduction}

Open-sourced and pre-trained models have helped to make tremendous progress in computer vision and machine learning research~\cite{golden_decade}. These open-source models improve the reproducibility of research and allow for fair comparisons across the models~\cite{sonnenburg2007need}. With the accessible pre-trained model such as image classifiers~\cite{he2016deep,sandler2018mobilenetv2,huang2017densely} trained on ImageNet~\cite{deng2009imagenet}, much research spurred out of these backbones building on top of them, \eg, detection~\cite{lin2017focal,redmon2017yolo9000}, segmentation~\cite{long2015fully,he2017mask}, and many other applications based on transfer learning~\cite{huh2016makes,kornblith2019better}. However, as computer vision and fine-tuning methods improve, open-sourced model weights may become a double-edged sword. With the ability to quickly fine-tune a model to a new task with few training samples, the entry barrier to developing a working computer vision system on a new task is greatly lowered; this includes bad actors developing potentially harmful applications, \eg, the ability to quickly train models on human subjects which may raise privacy concerns~\cite{payton2023privacy,zheng2024imma}.
In this work, we ask the following question:
\vspace{-0.2cm}
\begin{quote}
\textit{Is possible to develop a pre-trained model that is difficult to fine-tune for certain downstream tasks?} 
\vspace{-0.2cm}
\end{quote}
If we succeed, the pre-trained models can be released to support scientific research while addressing safety concerns. To begin this endeavor, we focus on the task of few-shot classification (FSC). We investigate whether it is possible to have a pre-trained model that FSC performs poorly on a set of restricted classes while remaining competitive on the remaining classes. We selected FSC to study as it is a well-established area with proper benchmark and fine-tuning procedures~\cite{fei2006one,snell2017prototypical,hu2022pmf,zhang2022tip,gao2023clip,zhou2022learning,song2022clip}  using pre-trained ImageNet backbones, \eg, ResNet~\cite{he2016deep}, and more recently on large-scale language and vision backbone, \eg, CLIP~\cite{radford2021learning}.

To achieve this goal, we propose Learning To Obstruct (\method), a MAML~\cite{finn2017model}-like algorithm, that learns a \textit{poor initilization} \wrt an FSC algorithm for the set of restricted classes.
We evaluate the proposed \method algorithm by conducting experiments on two few-shot classification setups: (a) the classic N-way-K-shot setting using ProtoNet~\cite{snell2017prototypical} and MetaOptNet~\cite{lee2019meta}; (b) the more recent language-vision few-shot learning setting using CoOp~\cite{zhou2022learning} and Tip-Adapter~\cite{zhang2022tip}. On ImageNet~\cite{deng2009imagenet} and CIFAR100~\cite{krizhevsky2009learning}, we show that \method successfully obstructs the learning of FSC methods, achieving lowered accuracy on restricted classes and maintained competition accuracy on other classes.
Lastly, we also experimented with applying LTO for attribute learning on the CelebA dataset~\cite{liu2015faceattributes}.

{\noindent\bf Our contributions are as follows:}
\begin{itemize}[topsep=3pt]
\item We propose the task of learning to obstruct FSC from learning in restricted classes.
\item We present \method, a meta-learning algorithm, that learns poor backbone initialization for obstructing FSC methods.
\item We conduct extensive experiments validating the effectiveness of \method on four different FSC algorithms on ImageNet, CIFAR100, and CelebA datasets.
\end{itemize}

\input{figs/intro_comp}

%% file: figs/intro_comp.tex
\begin{figure*}[t]
\centering
\includegraphics[width=0.995\linewidth]{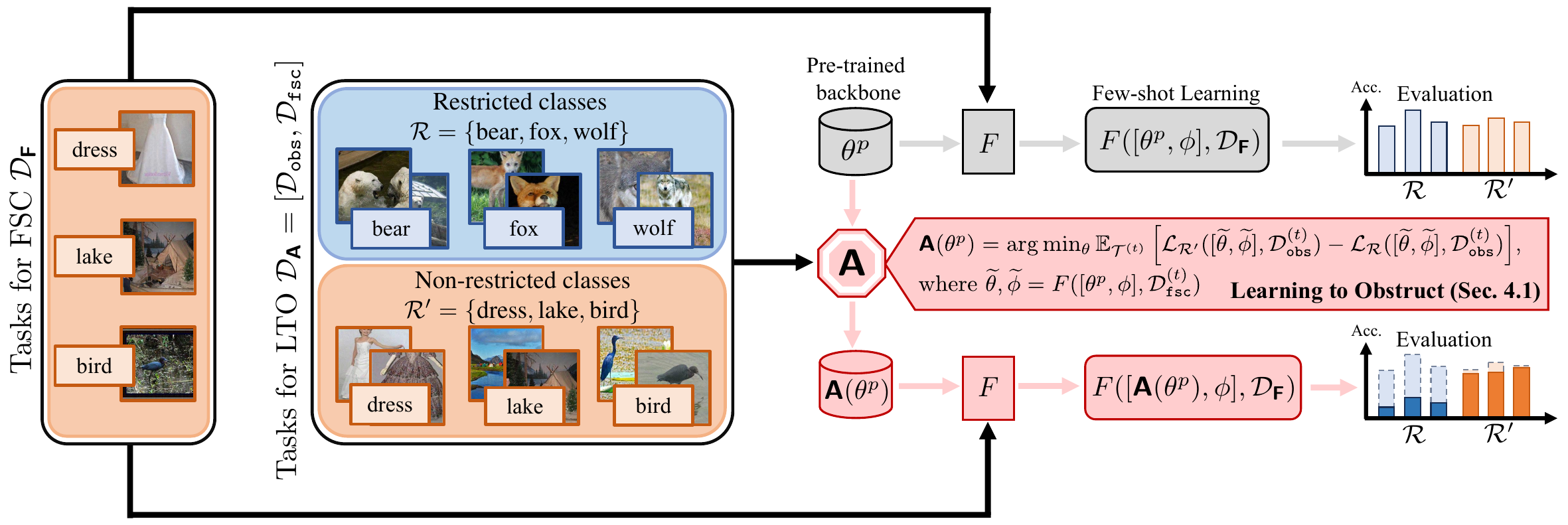}
\vspace{-0.35cm}
\caption{\textbf{Learning to Obstruct (LTO) few-shot learning paradigm.} \textit{\hlc[wolto]{Without LTO}}: after the adaptation of few-shot learner $F$, the model can classify classes from $\gR$ and $\gR'$ correctly. \textit{\hlc[OursColor]{With LTO}}: By modifying the pre-trained model parameters $\theta^p$ via our proposed method $\tA$ before the adaptation of $F$, the model fails to generalize to restricted class set $\gR$ while maintaining its performance in other class set $\gR'$. %
}
\vspace{-0.3cm}
\label{fig:intro_comp}
\end{figure*}

%% file: src/rel.tex
\section{Related Work}
\label{sec:related}

{\noindent\bf Few-shot learning.} %
Learning from limited labeled data is a crucial task in computer vision, which is known as few-shot learning (FSL)~\cite{fei2006one}. The mainstream of FSL can be categorized into three main branches: metric-based~\cite{snell2017prototypical,oreshkin2018tadam,yoon2019tapnet,qiao2018few,hao2019collect,vinyals2016matching}, optimization-based~\cite{finn2017model,jamal2019task,frikha2021few,rajeswaran2019meta,ravi2016optimization}, and augmentation-based methods~\cite{chen2019image,hariharan2017low,schwartz2018delta,wang2018low}. In this work, we focus on few-shot image classification~\cite{sung2018learning,chen2019closer,kang2021relational,wertheimer2021few,yang2022few,chen2021meta}, \ie, 
FSL but for image classification.

With the availability of pre-trained backbones, FSC methods now use pre-trained weights for initialization~\cite{phan2016differential}. The choice of pre-trained weights varies between the methods. For example, a common choice is pre-training on large-scale image classification tasks~\cite{deng2009imagenet,he2016deep}. More recently, the use of large-scale language and vision backbones have also been explored. For example, CLIP~\cite{radford2021learning} and DeCLIP~\cite{li2021supervision} have demonstrated remarkable capabilities in learning zero-shot transferable features across diverse datasets and domains. 
Building on top of these pre-trained backbones, CoOp~\cite{zhou2022learning}, CLIP-Adapter~\cite{gao2023clip}, and Tip-Adapter~\cite{zhang2022tip} show that the model performance can be further improved by prompt optimization~\cite{zhou2022learning}, fine-tuning the introduced light-layer ``Adapter''~\cite{gao2023clip} or even training-free adaptation~\cite{zhang2022tip}.
Without any restriction on the learnable classes, one can easily apply these few-shot methods on pre-trained backbones, even for classes with potential harm.

\myparagraph{Machine unlearning.} 
In machine unlearning, a model is trained to forget specific class(es) of data while retaining the memory of the rest without re-training from scratch. This concept is encapsulated in a data-forgetting algorithm proposed by \citet{cao2015towards}. For example, \citet{golatkar2021mixed} trained two separate networks: the core model and a mixed-linear model for unlearning purposes. \citet{tarun2023fast} introduced an error-maximization-based method to learn a noise matrix for the class to be forgotten. Recently, with the pre-trained language and vision backbone, machine unlearning is also used for de-biasing by forgetting certain attributes of the images. \citet{wang2021gender} removed the dimensions of CLIP embeddings that are highly correlated with the target attributes. %
\citet{foster2023fast} propose SSD, \ie selective synaptic
dampening, a swift and effective retrain-free method for machine unlearning, utilizing a two-step process to identify and diminish crucial parameters without the need for long-term storage of training data.
We note that the proposed task of obstructing FSC is not machine unlearning. In machine unlearning, the goal is the ``removal of'' certain classes, whereas our goal is the ``prevention of learning'' certain classes.

\myparagraph{Data poisoning attacks.} As the proposed \method can be viewed as ``ruining'' the backbone for the restricted classes, another potential way to accomplish that is through data poisoning attacks. Specifically, data poisoning attacks make changes to the training data to corrupt models' test-time behavior~\cite{xiao2015feature,koh2017understanding,schwarzschild2021just,shafahi2018poison,
dai2019backdoor}. 
In the context of few-shot learning,~\citet{oldewage2022adversarial} poisoned the support data in meta-testing, and achieved misclassification for the query prediction. However, this poisoning approach falls short of offering genuine protection for specific restricted classes. Individuals could still collect a few clean images for FSC. In contrast to these approaches, our proposed method obstructs the pre-trained backbone against the restricted classes.

%% file: src/prelim.tex
\section{Preliminaries}
\label{sec:prelim}
As our proposed learning to obstruct can be thought of as learning an initialization for few-shot classification, we provide a brief review of how to learn a good initialization, specifically on MAML~\cite{finn2017model}, followed by the background of few-shot classification~\cite{vinyals2016matching,chen2021meta,hu2022pmf}.

\myparagraph{Learning a good initialization.}
At a high level, MAML is an algorithm that aims to learn a ``good initialization'' for a model being trained on a new task using gradient-based methods. MAML assumes a distribution of tasks $P(\gT)$ where each task $\gT^{(t)} = (\gS^{(t)}, \gQ^{(t)}) \sim P(\gT)$ is comprised of a support set $\gS$ and a query set $\gQ$. 

The task of learning a good initialization is then formulated as the following optimization problem:
\bea\label{eq: maml_target}
   \vartheta^\star = \arg \min_{\vartheta} \sE_{\gT^{(t)} \sim P(\gT)}\left[ \gL^M\left(U(\vartheta, \gS^{(t)}), \gQ^{(t)} \right) \right],
\eea
where $\gL^M$ denotes a loss for the task in MAML, and $U$ denotes a learner function that updates the model parameter $\vartheta \in \sR^d$, \eg, MAML chooses $U$ to be a single gradient update. Intuitively, $\gL$ is evaluating the quality of $\vartheta$ when used as initialization to the learner $U$. That is, MAML is searching for a good initialization such that after applying the learner $U$ the model performs well.

To solve the optimization problem in~\equref{eq: maml_target}, MAML uses gradient descent and approximates the gradient only using first-order terms. 
In summary, the optimization problem in~\equref{eq: maml_target} can be decomposed into two parts: \textbf{(a)} a learner function $U$ that modifies the model parameters, %
and \textbf{(b)} an outer optimization that minimizes $\gL$ usually solved using gradient descent for deep nets. This general framework serves as the basis for our proposed \method algorithm.

\myparagraph{Few-shot classification (FSC).}
The goal of few-shot classification is to train a model such that it generalizes well to novel tasks $\gT^{(T+1:T+M)} \sim P(\gT)$ given a dataset of training tasks $\gT^{(1:T)} \sim P(\gT)$. For few-shot classification, the support set $\gS= \{(\vx_s,\vy_s)\}$ and query set $\gQ= \{(\vx_q,\vy_q)\}$ contains input images $\vx$ with the corresponding groundtruth $\vy$ in the class space $\gY^{(t)}$ of each task. Typically, the task follows a $N$-way-$K$-shot setup, \ie, each support set $\gS$ contains $N$ classes and $K$ examples per class.

For most FSC methods~\cite{snell2017prototypical,vinyals2016matching,sung2018learning}, 
a prediction $\hat{\vy}_q$  given $\vx_q$ is made using a predictor $\hat F$, \ie,
$
\hat{\vy}_q = \hat{F}(\vx_q, \vartheta, \gS^{(t)}),
$
where the prediction depends on the examples in the support set $\gS^{(t)}$. Note, $\hat{\vy}_q$ is a vector of the predicted probability of each class.  
To train this model, FSC methods train the model using all the training tasks $\gT^{(1:T)} = \{\gT^{(1)}, \dots, \gT^{(T)}\}$, \ie, 
\bea\label{eq: fsl_loss}
&\gL^F(\vartheta, \gT^{(1:T)}) = \sum\limits_{\gQ^{(t)} \in \gT^{(1:T)}}\sum\limits_{ (\vx_q, \vy_q) \in \gQ^{(t)}} \ell (\vy_q, \hat \vy_q) 
\eea
with $\ell$ denoting the cross-entropy loss on a single sample.
Optimizing~\equref{eq: fsl_loss} gives rise to a learner function %
\bea\label{eq: fsl_update}
F(\vartheta, \gT^{(1:T)}) = \arg \min_{\vartheta} \gL^F(\vartheta, \gT^{(1:T)}),
\eea
which outputs the optimal model parameters $\widetilde \vartheta$ given the set of training tasks $\gT^{(1:T)}$.
In summary, a FSC algorithm $\tF = (\hat{F}, F)$ is composed of a $\hat F$ yielding the prediction for each query sample and $F$ learner function that updates the model parameters towards the solution that minimizes the loss~\equref{eq: fsl_loss}.

For  FSC method $\tF$ using deep-nets, the model parameter $\vartheta=[\theta, \phi] \in \sR^{d}$ is further decomposed into two parts: $\theta$ for the parameters of the backbone $g_{\theta}$ and $\phi$ for the parameters of the classifier $f_{\phi}$ used for constructing the prediction function $\hat{F}$. The choice of $g_\theta$ and $f_\phi$ depends on the FSC method. %
For example, ProtoNet~\cite{snell2017prototypical} chooses a classifier $f_{\phi}$ to be the normalized distance of an input query's feature to the prototypes,~\ie,
$ %
\hat{F}(\vx_q, [\theta, \phi], \gS)[k] =  \frac{\exp(-d(g_\theta(\vx_q), \vr_k))}{
\sum_{k'} \exp(-d(g_\theta(\vx_q), \vr_k'))
}
$
where $d$ corresponds to a distance function with prototypes
$\vr_k = \frac{1}{\gS_k} \sum_{(\vx_s, \vy_s) \in \gS} g_\theta(\vx_s)
$ %
defined as the average of samples in the support set with the label $k$ denoted as $\gS_k$. %

Next, to further improve model performance, instead of training $\theta$ from scratch, recent works~\cite{chen2021meta, hu2022pmf,chen2019closer} introduce pre-trained backbones $\theta^p$ as the initialization to $\theta$ when the training a few-shot classifier.

\myparagraph{Language and vision based FSC.} 
CLIP~\cite{radford2021learning} is a powerful foundation model that encodes rich language and vision information. It serves as a strong pre-trained backbone $\theta^p$ for few-shot or even zero-shot learning.
CLIP consists of a text encoder and a visual encoder, i.e., $\theta^p = [\theta^p_{\texttt{text}}, \theta^p_{\texttt{img}}]$.
To build a few-shot classifier~\cite{gao2023clip, zhang2022tip, sung2022vl, zhou2022conditional, zhou2022learning} over classes $\gY$ with CLIP, the predictor $\hat{F}$ is defined as follows:
$
\hat{F}(\vx_q, [\theta, \phi], \gS)[k] = \frac{ \exp(\vv_{k}^{\intercal}\vv_{\vx_q})
}{\sum_{k'} 
\exp( \vv_{k' \in \gY}^{\intercal}\vv_{\vx_q})},
$
where $\vv_\vx = g_{[\theta_{\tt img},\phi]}(\vx)$ and $\vv_k = g_{[\theta_{\tt text},\phi]}(k)$ corresponds to image and class features extracted from the encoders $g$. We note that there are ``implicit'' dependencies on the support set $\gS$, as the encoders $g$ are fine-tuned on $\gS$ in this language and vision-based FSC methods.

%% file: src/app.tex
\section{Approach}
In this paper, our goal is to \textit{obstruct} the learning of specific classes in a restricted class set $\gR$, when utilizing few-shot classification (FSC) methods. At the same time, we aim to ensure that the model's performance in the other class set $\gR'$ remains unaffected. We consider the scenario where the FSC algorithms $\tF = (\hat F, F)$ using a pre-trained backbone and are known to the obstructor. To achieve this, we introduce the Learning To Obstruct (\method) algorithm $\tA$ that modifies in the pre-trained backbone's parameters $\theta^p$ \textit{to create a poor initialization} $\tA(\theta^p)$. When the FSC algorithm is applied to $\tA(\theta^p)$, the model will perform poorly on 
restricted classes but not the other classes. 

\subsection{Learning to obstruct}
{\noindent\bf Problem formulation.} As the name suggests, \method is formulated as a learning problem. In this learning problem, we are given a distribution of tasks $P(\gT)$ for which we can sample tasks $\gT^{(t)}$ containing the support and query set $(\gS^{(t)}, \gQ^{(t)})$, and a set of restricted classes $\gR$. We further define the set of ``other classes'' as $\gR' = \{k \in \gY : k \notin \gR\}$, where $\gY = \bigcup_t \gY^{(t)}$ denotes the set of all possible classes across all tasks.

For our algorithm, we further split the data into two parts,
\bea
\gD^{(t)}_{\tt obs} = (\gS^{(t)}_{\tt obs}, \gQ^{(t)}_{\tt obs}) \text{~~and~~} \gD^{(t)}_{\tt fsc} = (\gS^{(t)}_{\tt fsc}, \gQ^{(t)}_{\tt fsc}). 
\label{eq:D_split}
\eea
The $\gD^{(t)}_{\tt obs}$ split is for evaluating the quality of the obstruction, and  $\gD^{(t)}_{\tt fsc}$ is to be used for training by the FSC learner function $F$. 
Using these splits, we formulate the Learning To Obstruct (\method) algorithm $\tA$ as an optimization problem:
\bea
\label{eq: main_target}
    &\min\limits_{\theta} \sE_{\gT^{(t)}} \left[\gL_{\gR'}\left([\widetilde \theta, \widetilde \phi], \gD^{(t)}_{\tt obs}\right) - \gL_{\gR}\left([\widetilde \theta, \widetilde \phi], \gD^{(t)}_{\tt obs}\right) \right]
\text{~s.t.~} \widetilde \theta, \widetilde \phi = F([\theta, \phi], \gD^{(t)}_{\tt fsc}).
\eea
The objective in~\equref{eq: main_target} consists of two terms $\gL_{\gR'}$ and $\gL_{\gR}$ which corresponds to the few-shot learning loss $\gL^F$ in~\equref{eq: fsl_loss} but evaluated only on \textit{other classes} $\gR'$ and restricted classes $\gR$ respectively for the query set. Formally,
\bea\nonumber
\gL_\gR([\widetilde \theta, \widetilde \phi], \gD^{(t)}_{obs}) = \gL^F\Big([\widetilde \theta, \widetilde \phi],
{\color{myblue}(}\gS_{\tt obs}^{(t)}, \{(\vx_q, \vy_q) \in \gQ^{(t)} \land \vy_q \in \gR\}{\color{myblue})}\Big)
\eea
and vice versa for $\gR'$.
Intuitively~\equref{eq: main_target}, inspired by MAML, aims to learn a poor initialization of model weights for classes in $\gR$, we first let $\theta$ explore the landscape
\begin{wrapfigure}[17]{r}{0.7\linewidth}%
\vspace{-1.6cm}
\begin{minipage}{0.7\textwidth}%
\input{algs/lto}%
\end{minipage}%
\end{wrapfigure}%
 using the FSC's learner $F$, then we collect the gradients by maximizing $\gL_{\gR}$ and at the same time minimize $\gL_{\gR'}$.\\ 
\indent\myparagraph{Optimization.}
To solve the optimization problem in~\equref{eq: main_target}, we use a gradient-based method. We illustrate the overall algorithm in~\algref{alg:lto} using mini-batch gradient descent. Given a randomly sampled batch of tasks $\gB$, the FSC learner $F$ updates the model parameters based on $\gD^{(t)}_{\tt fsc}$ separately for each task to produce updated parameters $\widetilde{\theta}$ and $\widetilde \phi$. To backpropagate through, the learner defined in~\equref{eq: fsl_update}, we approximate the $\argmin$ with unrolled gradients. %

Next, based on whether the example is of a class belonging to $\gR$, we compute $\gL_{\gR}$ and $\gL_{\gR'}$ over $\gD^{(t)}_{\tt obs}$ with updated parameters $\widetilde \theta$ and $\widetilde \phi$.
We then compute the gradient $\Delta \theta^{(t)}$ with bacpropgation on $\gL_{\gR'} - \gL_{\gR}$ \wrt $\theta$. After collecting all the $\Delta \theta^{(t)}$ within one batch, we update $\theta$ with the aggregated gradient. This training procedure allows the model parameters to be steered to an unfavored spot conditioned on $\gR$ while maintaining or even enhancing its performance of generalizing to unrestricted classes after applying a downstream few-shot learner.

\myparagraph{From classification to multi-label classification.} 
Beyond the classification formulation, \method can also be formulated to obstruct attribute learning. We treat attribute learning as a multi-label classification problem, \ie, each input $\vx$ is labeled with a vector label $\vk \in \sN^{|\gA|}$, where $\gA$ denotes the set of all attributes. For each attribute $a \in \gA$, an algorithm $\tF$ builds a designated predictor and introduces additional parameters $\phi_a$ to build a model with parameters $\vartheta = [\theta, \phi_{1:|\gA|}]$.

In attribute learning, LTO aims to obstruct a set of restricted attributes $\gR \subset \gA$. The algorithm follows~\algref{alg:lto}, except we change the objective in~\equref{eq: main_target} to the following:
\bea
\gL_\gR([\widetilde \theta, \widetilde \phi_{1:|\gA|}], \gD^{(t)}_{obs}) = \sum_{a \in \gR} \gL^{F}\Big([\widetilde \theta, \widetilde \phi_a], \gD^{(t)}_{obs} \Big),
\eea and vice versa for $L_{\gR'}$.

\subsection{Additional Details}

\myparagraph{Gradient computation for $\theta$.}
Implementation-wise, to collect all $\Delta \theta^{(t)}$ for each task $\gT^{(t)}$ within one batch with the same initial parameters, we restore $[\theta, \phi]$ to the values at the beginning of each epoch. 
That is, after collecting all $\Delta \theta^{(t)}$s and updating $[\theta, \phi]$ at the end of the epoch, we cache the updated parameters for future restoration. 

\myparagraph{Resampling prompts and texts for CLIP-based FSC.}
In CLIP-baed FSC, a prompt set $\gP$ contains text templates to guide the output of the text encoder. Each prompt $\vp \in \gP$ usually takes the form of ``a photo of \{k\}'' in image classification, where $k \in \gY$ is a class label. 
The classifiers in CLIP-based FSC are usually built on the text features of $\gP$ and $\gY$. That is, for each class label $k \in \gY$, the class feature
\bea
\vv_k = g_{[\theta_\texttt{text}, \phi]}(k) = \texttt{Agg}_{\vp \in \gP} g_{[\theta_\texttt{text}]}(\vp(k)),
\eea where $\texttt{Agg}(\cdot)$ is an aggregaion function.

Due to the limitation of GPU memory, we do not compute the gradients of text features generated from all prompts $\gP$ and all classes $\gY$. Instead, for every few steps, we use an unbiased estimate by
re-sampling a subset of prompts $\gP' \subset \gP$ and classes $\gY' \subset \gY$. 
For each $k \in \gY'$, its correspondent $\vv_k$ is defined as $\texttt{Agg}_{\vp \in \gP'} g_{[\theta_\texttt{text}]}(\vp(k))$,
and only the gradients from these tensors $\{\vv_k \mid k \in \gY'\}$
are backpropagated.

%% file: algs/lto.tex
\begin{algorithm}[H]
\caption{Learning to Obstruct (Our method)}
\begin{algorithmic}[1]
\renewcommand{\algorithmicrequire}{\textbf{Input:}}
\renewcommand{\algorithmicensure}{\textbf{Output:}}
\REQUIRE pre-trained backbone: $\theta^p$, %
task distribution: $P(\gT)$, epoch: $I$, learning rate: $\alpha$, restricted classes: $\gR$, FSC loss: $\gL^F$, few-shot learner: $F$.

\ENSURE  Obstructive backbone: $\theta$
\STATE Intialize $\theta = \theta^p$, and $\phi$ following the FSC method.
\FOR {$i = 1$ to $I$}
\STATE Sample batch $\gB: \{\gT^{(t)}\}_{t=1}^{|\gB|}\sim P(\gT)$,\\ where $\gT^{(t)} = (\gD^{(t)}_{\tt fsc}, \gD^{(t)}_{\tt obs})$
\FOR {$t = 1, \dots, |\gB|$}
\STATE $\widetilde{\theta}, \widetilde{\phi} = F([\theta, \phi], \gD^{(t)}_{\tt fsc})$
\STATE $\Delta \theta^{(t)} = \nabla_\theta \left[ \gL_{\gR'}\left([\widetilde{\theta}, \widetilde{\phi}], \gD^{(t)}_{\tt obs}\right) - \gL_{\gR}\left([\widetilde{\theta}, \widetilde{\phi}], \gD^{(t)}_{\tt obs}\right) \right]$
\ENDFOR
\STATE $\theta \leftarrow \theta - \alpha \sum_{t=1}^{|\gB|} \Delta \theta^{(t)}$
\ENDFOR
\RETURN $\theta$
\end{algorithmic}
\label{alg:lto}
\end{algorithm}

%% file: src/exp.tex
\input{figs/selection_r}

\section{Experiments}
We first describe how we construct the restricted class set followed by the dataset preparation in our experiments. Dataset-specific details are described subsequently.

\myparagraph{Selection of restricted classes $\gR$.}
To resemble real-world scenarios, choose the set of restricted classes including individual classes that are related to each other. Specifically, we divide $\gY$ into $N'$ superclasses $\{\gY_1, \cdots, \gY_{N'}\}$, where each $\gY_n$ consists of $k$ with similar semantics following existing work~\cite{robustness, krizhevsky2009learning}.
Once superclass $\gY_n$ is picked then all subsuming classes $k \in \gY_i$ are categorized into $\gR$ while the remaining classes are put into $\gR'$. \figref{fig:selection-r1} provides an illustration where we extract 4 superclasses for image classification. Similarly, in~\figref{fig:selection-r2}, we illustrate how the restricted classes are chosen for attribute learning.

\myparagraph{Training and evaluation data split.} To show the obstruction effect of LTO, we use FSC algorithm $\tF$ with backbone parameters initialized as $\tA(\theta^p)$. %
To avoid data/class information leakage in the experiment setup, carefully consider three disjoint dataset splits $\{\gD_\tA, \gD_{\tF}, \gD_{\texttt{eval}}\}$ each corresponds to data used by \method, by FSC in meta-training, and the evaluation set for meta-testing. We note that $\gD_\tA$ is used for sampling $\gD_{\texttt{fsc}}$ and $\gD_{\texttt{obs}}$ in~\equref{eq:D_split}, \ie, $\gD_\tA = \bigcup_{t=1}^{|\gB|} \gT^{(t)} = \bigcup_{t=1}^{|\gB|} (\gD_{\texttt{fsc}}^{(t)} \bigcup \gD_{\texttt{obs}}^{(t)})$.

\myparagraph{Baselines.}
We compare LTO to two baselines: %
\begin{itemize}[topsep=1pt]
    \item \bsla aims to ``ruin'' the backbone by directly maximizing the loss for the restricted set $\gR$,~\ie,
    \bea
    \max\limits_{\theta} \sE_{\gT^{(t)}} \left[\gL_{\gR}\left([\theta, \phi], \gT^{(t)} \right) \right].
    \eea
    \item \bslb chooses to minimize $\gL_{\gR'} - \gL_{\gR}$ 
    without the consideration of the FSC algorithm $\tF$. This is equivalent to removing the computation of $\widetilde \theta$ in LTO,~\ie,
    \bea
    \min\limits_{\theta} \sE_{\gT^{(t)}} \left[\gL_{\gR'}\left([\theta, \phi], \gT^{(t)} \right) - \gL_{\gR}\left([\theta, \phi], \gT^{(t)} \right)\right].
    \eea
\end{itemize}
For both baselines, for each FSC method, we use the same amount of data and hyperparameters as in \method for all superclasses.

\input{src/exp-01-classic}

\input{src/exp-02-clip}

\input{src/exp-03-attr}

\input{src/exp-05-unlearning}

%% file: figs/selection_r.tex
\begin{figure*}[t]
\begin{minipage}[t]{0.475\linewidth}
    \includegraphics[width=\linewidth]{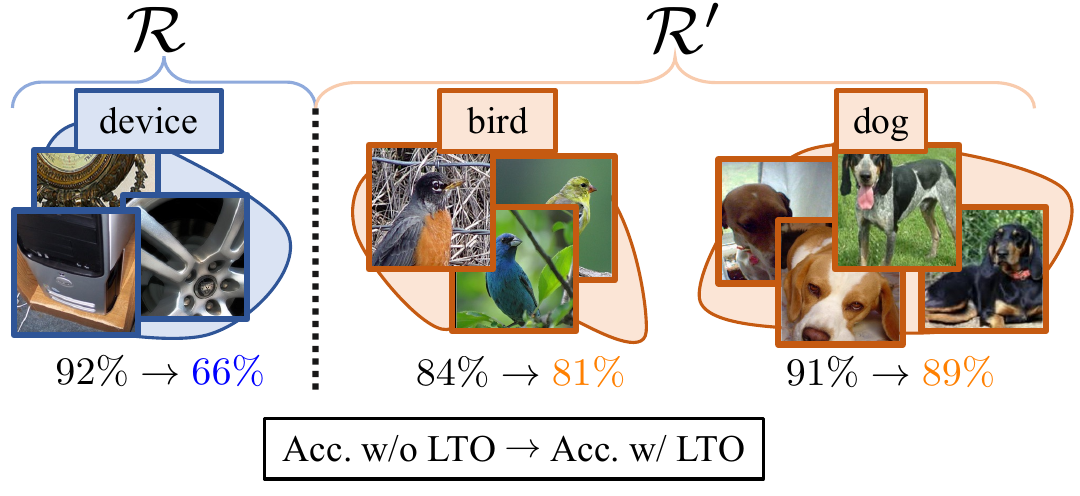}
    \vspace{-0.5cm}
    \caption{
        \textbf{Selection of $\gR$ on image classification.} 
        The objective of LTO on image classification is to minimize the top-1 Acc. on $\gR$ while maintaining the top-1 Acc. on $\gR'$. In this example, $\gR = \gY_{\text{device}}$
        while $\gR' = \gY_{\text{bird}} \bigcup \gY_{\text{dog}}$. LTO decreases the top-1 Acc. on $\gR$ from 92\% to 66\%, while the drops on $\gR'$ are no more than 3\%. 
    }
        \label{fig:selection-r1}
\end{minipage}%
\hfill
\begin{minipage}[t]{0.475\linewidth}
    \includegraphics[width=\linewidth]{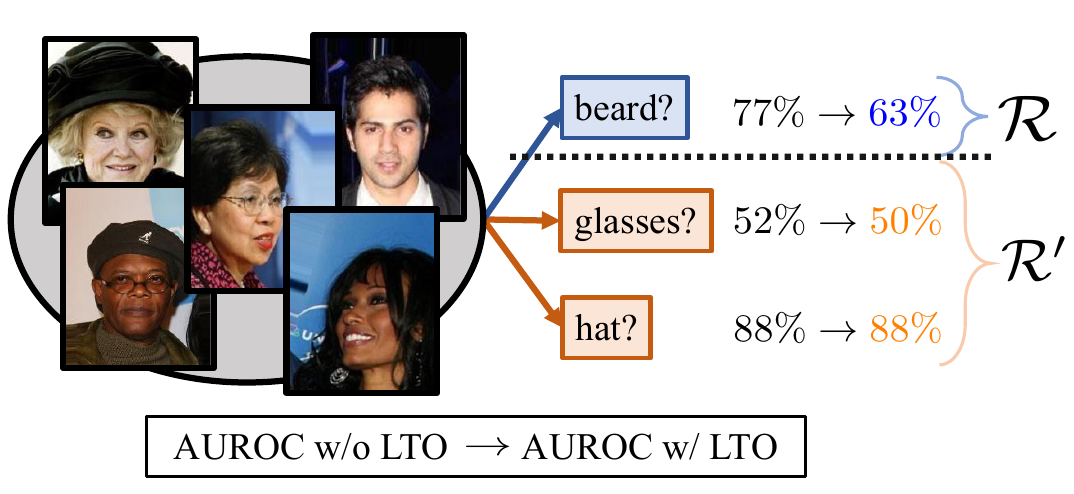}
    \vspace{-0.5cm}
    \caption{
        \textbf{Selection of $\gR$ on attribute learning.}
        The objective of LTO on attribute learning is to minimize the AUROC on $\gR$ while maintaining the performance on $\gR'$. In this example, $\gR = \{ \text{bald} \}$ while $\gR'=\{ \text{hat}, \text{glasses} \}$. LTO decreases the AUROC on $\gR$ from 77\% to 63\%, while the drops on $\gR'$ are no more than 2\%. 
    }
    \label{fig:selection-r2}
\end{minipage}
\hfill

\vspace{-0.5cm}
\end{figure*}

%% file: src/exp-01-classic.tex
\subsection{Obstruct Classical Few-shot Classification}
\label{sec:exp-obs-classifical}

We perform experiments by choosing $\tF$ to be classical FSC methods, including, ProtoNet~\cite{snell2017prototypical} and MetaOptNet~\cite{lee2019meta}. We demonstrate that by incorporating our LTO algorithm $\tA$, the performance of $\tF(\tA(\theta^p))$ in the restricted class set $\gR$ declines significantly while the performance in $\gR'$ is mostly maintained.

\myparagraph{Experiment setup.} Following the setting of~\citet{hu2022pmf}, we choose a pre-trained ResNet18~\cite{he2016deep} as $\theta^p$. 
For $\gR$ selection on ImageNet, we choose to group classes based on the superclasses provided by~\citet{robustness}.
\begin{wrapfigure}[16]{r}{0.37\linewidth}
    \centering
    \vspace{-.88cm}
    \includegraphics[width=\linewidth]{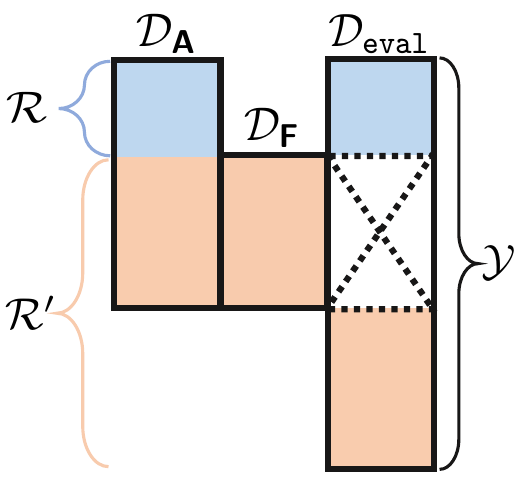}
    \vspace{-0.65cm}
    \captionof{figure}{
        \textbf{Data split for LTO on classical FSC.} For  LTO on classical FSC, we split the dataset into three disjoint sets $\gD_\tA$, $\gD_\tF$, and $\gD_{\texttt{eval}}$. The set $\gD_\tA$ is for LTO; the set $\gD_\tF$ is for training $\tF$ after LTO; $\gD_{\tt eval}$ is for evaluation. %
    }
    \label{fig:fs_data_intro}
\end{wrapfigure}
In total, there are 10 superclasses and each superclass subsumes 38 of the original ImageNet classes. For each experiment, we select one out of ten superclasses as $\gR$.
We use the split of the first superclass ($\text{id}=0$) as the validation task for tuning hyperparameters. We report the result for the rest of the superclasses from $\text{id}=1 \text{ to } 9$. Additional details of the superclass are provided in the appendix.
For data split, we visualize it in~\figref{fig:fs_data_intro}. Since we are evaluating the performance of $\tF$ in novel classes, both $\gR$ and $\gR'$ in $\gD_{\tt eval}$ should not show up in $\gD_\tF$. Thus, We further split $\gR'$, with 70\% classes for $\gD_{\tF}$ and 30\% for $\gD_{\tt eval}$.\\
\indent\myparagraph{Training details.}
For all of the experiments, we ran 200 steps of obstructive learning with a batch size of 20. Within each obstructive step, we ran 20 gradient steps for the FSC learner $F$.\\ 
\indent\myparagraph{Evaluation metric.}
We assess the quality of obstruction by measuring the \textit{gap in} top-1 classification accuracy with and without \method. We denote the decrease in top-1 Acc. of FSC when using \method for the restricted class $\gR$ and the other classes $\gR'$ as $\delta_{\gR}$ and $\delta_{\gR'}$ respectively. 

For comparison, we propose the evaluation metric
\texttt{DropRatio}@$\beta$ denoted as
$\Delta@\beta = \frac{\delta_\gR}{\delta_{\gR'}}$,
which corresponds to the ratio of the accuracy drop of the restricted classes over the other classes. We select the model when the accuracy drop on the other classes $\delta_{\gR'}$ is closest to $\beta$\%. The higher the drop ratio, the better the obstruction.\\
\vspace{-.3cm}
\begin{wrapfigure}[13]{l}{0.5\linewidth}%
\vspace{-1cm}
\input{figs/fs_acc}
\end{wrapfigure}%
\myparagraph{Results.}
In~\figref{fig:fs_acc}, we visualize the average accuracy vs. the obstructive learning steps on all nine superclasses of ImageNet with and without LTO on the selected classical FSC methods. The gap between the solid and dashed lines illustrates the accuracy drop due to \method, while the blue and orange lines highlight the restricted and other classes respectively. As can be observed, the gaps are consistently observed across all the plots.
For MetaOptNet, we do not observe 2\% drop in $\gR'$ within 200 steps, hence we selected the model of the last step. 

\begin{wrapfigure}[19]{r}{0.5\textwidth}
\vspace{-.35cm}

\input{tables/table_main_fs}

\end{wrapfigure}
Next, we report the proposed evaluation metric of \texttt{DropRatio} with $\beta=2$, \ie, $\Delta@2$  when obstructed using the baselines and our~\method for comparison. Intuitively, this corresponds to the ratio between the gap (between orange and blue lines) in accuracy at the vertical line shown in~\figref{fig:fs_acc}. Results are shown in~\tabref{tab:main_fs}. Higher the \texttt{DropRatio} indicates stronger obstruction on the FSC method.
For the column of \method \textit{\hlc[OursColor]{highlighted in pink}}, we observe that on average $\Delta@2$ is much larger than 1 under all settings, which means that our method can %
obstruct $\gR$ effectively without ruining the accuracy in $\gR'$. %
Overall, our method worsens the model's performance in the restricted classes while maintaining the accuracy of other classes.

%% file: figs/fs_acc.tex
\centering
\includegraphics[width=0.48\linewidth]{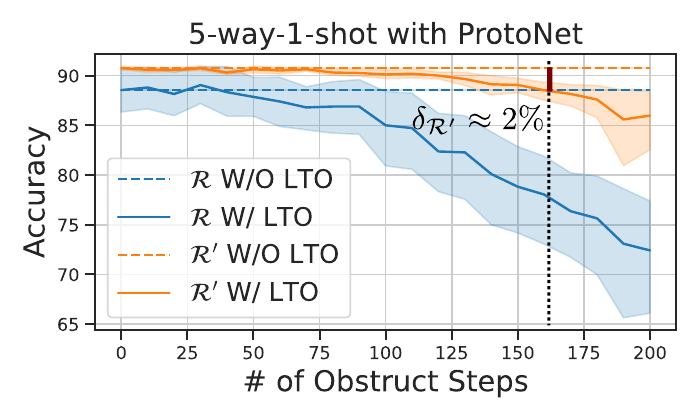}
\includegraphics[width=0.48\linewidth]{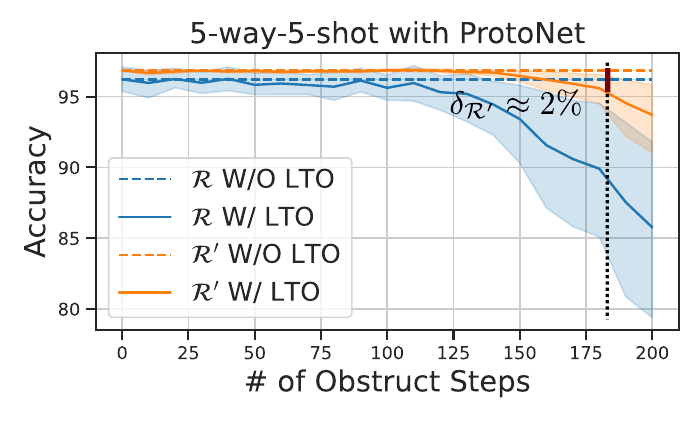}\\
\includegraphics[width=0.48\linewidth]{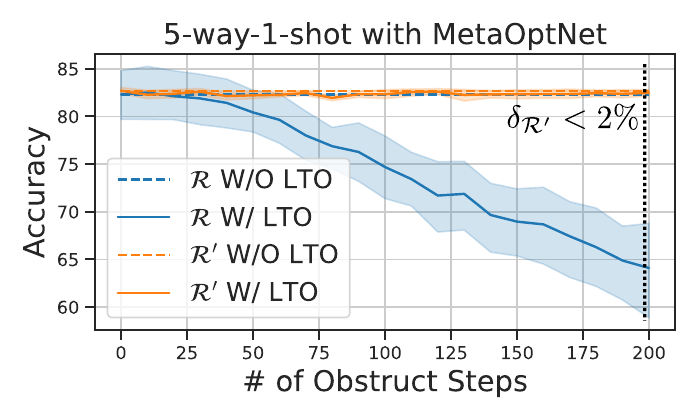}
\includegraphics[width=0.48\linewidth]{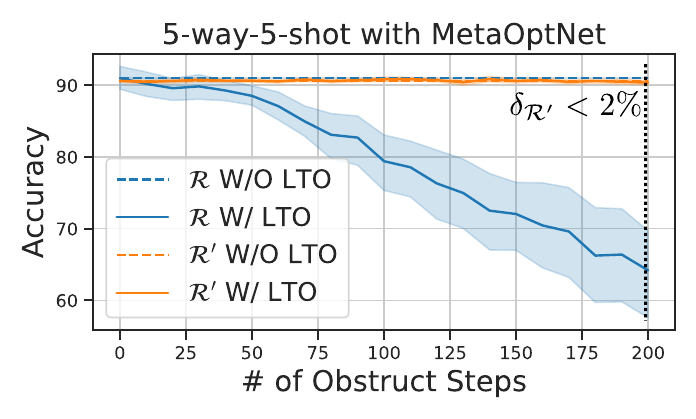}
\vspace{-0.25cm}
\captionof{figure}{\textbf{Accuracy (\%) of classical few-shot learning on ImageNet.} Gaps between different line types show the effect of LTO. Gaps between different colors show the effect on $\gR$ and $\gR'$.%
}
\vspace{-0.25cm}
\label{fig:fs_acc}

%% file: tables/table_main_fs.tex
\captionof{table}{\textbf{\texttt{DropRatio} of LTO on classical few-shot learning.} We report $\Delta@2$ on ImageNet over 9 selected superclasses. All experiments are 5-way classification.
}
\label{tab:main_fs}
\centering
\setlength{\tabcolsep}{4pt}
\resizebox{1.01\linewidth}{!}{
\begin{tabular}{
    c@{\hskip 1pt}
    c@{\hskip 2pt}
    rr
    >{\columncolor{OursColor}}r
    c
    rr
    >{\columncolor{OursColor}}r
}
\specialrule{.15em}{.05em}{.05em}
\multirow{2}{*}{$\tF$} & 
\multirow{2}{*}{$\gR$} & 
\multicolumn{3}{c}{1-shot} && 
\multicolumn{3}{c}{5-shot} \\
& & 
\bsla & \bslb & Ours &&
\bsla & \bslb & Ours \\ 
\cmidrule(lr){1-9}
\multirow{10}{*}{\rotatebox[origin=c]{90}{ProtoNet~\cite{snell2017prototypical,hu2022pmf}}}
& 1 & 1.07 & 2.21 & \bf 7.85 & ~ & 0.93 & 0.96 & \bf 1.69 \\ 
& 2 & 1.22 & 3.38 & \bf 3.93 & ~ & 1.15 & 2.31 & \bf 2.63 \\ 
& 3 & 1.05 & 2.15 & \bf 3.21 & ~ & 0.94 & \bf 1.98 & 0.05 \\ 
& 4 & 1.09 & \bf 6.91 & 6.60 & ~ & 1.36 & 2.78 & \bf 2.79 \\ 
& 5 & 1.37 & \bf 2.54 & 1.89 & ~ & 1.25 & 1.39 & 3.19 \\ 
& 6 & 1.05 & 4.82 & \bf 6.62 & ~ & 1.33 & 2.59 & \bf 2.90 \\ 
& 7 & 1.05 & 2.96 & \bf 4.30 & ~ & 1.08 & 2.33 & \bf 4.51 \\ 
& 8 & 1.00 & \bf 4.62 & 2.59 & ~ & 0.84 & 0.89 & \bf 3.16 \\ 
& 9 & 1.03 & \bf 4.28 & 2.80 & ~ & 1.02 & \bf 2.82 & 0.69 \\ 
\cmidrule(lr){2-9}
& Avg. & 1.10 & 3.77 & \bf 4.42 & ~ & 1.10 & 2.00 & \bf 2.40
\\
\cmidrule(lr){1-9}
\multirow{10}{*}{\rotatebox[origin=c]{90}{MetaOptNet~\cite{lee2019meta,hu2022pmf}}}
& 1 & 2.35 & \bf 12.06 & 8.71 & ~ & 2.64 & \bf 16.45 & 14.08 \\ 
& 2 & 2.21 & \bf 7.14 & 6.41 & ~ & 2.02 & 6.99 & \bf 11.32 \\ 
& 3 & 2.08 & 3.53 & \bf 5.22 & ~ & 1.74 & 5.31 & \bf 7.84 \\ 
& 4 & 2.52 & \bf 12.05 & 9.42 & ~ & 2.39 & 14.10 & \bf 14.98 \\ 
& 5 & 1.56 & 4.94 & \bf 6.84 & ~ & 1.53 & 7.34 & \bf 10.07 \\ 
& 6 & 1.57 & 14.54 & \bf 15.11 & ~ & 2.06 & 13.38 & \bf 22.44 \\ 
& 7 & 1.60 & 8.88 & \bf 9.61 & ~ & 1.91 & 9.77 & \bf 13.94 \\ 
& 8 & 2.27 & 13.53 & \bf 17.01 & ~ & 1.51 & 12.87 & \bf 21.57 \\ 
& 9 & \bf 1.37 & 1.16 & 1.34 & ~ & 1.63 & \bf 4.79 & 4.34 \\ 
\cmidrule(lr){2-9}
& Avg. & 1.95 & 8.65 & \bf 8.85 & ~ & 1.94 & 10.11 & \bf 13.40 \\ 
\specialrule{.15em}{.05em}{.05em}
\end{tabular}
}

%% file: src/exp-02-clip.tex
\subsection{Obstruct CLIP-based Few-shot Classification.}
\label{sec:exp_clip}

We perform experiments using CLIP-based few-shot algorithms, including CoOp~\cite{zhou2022learning}, TipAdapter~\cite{zhang2022tip}, and a simple baseline method,  named Cross-Entropy Fine-Tune(CE), which optimizes the CLIP's parameters by minimizing cross-entropy loss. We demonstrate that LTO can also obstruct foundation models with rich knowledge of both language and vision.

\myparagraph{Experiment setup.}
We select CLIP-ResNet50~\cite{radford2021learning} as the pre-trained backbone. 
We use ImageNet~\cite{deng2009imagenet} and CIFAR100~\cite{krizhevsky2009learning} to evaluate the performance of our LTO on clip-based image classification.
For restricted classes $\gR$ selection on CIFAR100, we follow the superclass groups provided by~\citet{krizhevsky2009learning}. The 100 classes in CIFAR100 are categorized into 20 superclasses, each containing five classes. For each experiment, we select one of the 20 superclasses as $\gR$ while the rest as $\gR'$.
The first superclass ($\text{id}=0$) is reserved as the validation task for hyperparameter tuning.
For $\gR$ selection on ImageNet is the same as in~\secref{sec:exp-obs-classifical}.

Unlike classical FSC methods,
CLIP-based FSC methods~\cite{zhang2022tip, zhou2022learning} focus on learning for one single task,~\ie, the $|\gY|$-way $K$-shot support set with the entire testing split as the query set.
Hence, $\gD_\tA$, $\gD_\tF$, and $\gD_{{\texttt{eval}}}$ all have access to every classes in $\gR$ and $\gR'$.
Additionally, since few-shot learning is motivated by the scarcity of high-quality labeled data, affecting both LTO and FSC for evaluation, we set $\gD_\tA$ to have the same few-shot setting as $\gD_{\tF}$. 
Specifically, for both ImageNet and CIFAR100, we sample a $|\gY|$-way 5-shot set from the training split as $\gD_\tA$ and another $|\gY|$-way 5-shot set as $\gD_{\tF}$. 
For each batch task $\gT^{(t)}$, we sample a $|\gY|$-way 1-shot set from $\gD_\tA$ as $\gD^{(t)}_{\texttt{fsc}}$ and a $|\gY|$-way 4-shot set as $\gD^{(t)}_{\texttt{obs}}$.\\ 
\begin{wrapfigure}[27]{r}{0.5\linewidth}
\vspace{-.4cm}
\input{tables/clip_main}
\end{wrapfigure}
\indent\myparagraph{Training details.} 
For CoOp~\cite{zhou2022learning} and TipAdapter~\cite{zhang2022tip}, we follow their training hyperparameters when used for evaluating LTO. Additional training details are in the appendix.\\
\indent\myparagraph{Results.} We report on combinations of LTO methods and few-shot algorithms $\tF$. Each $\tA(\theta^p)$ under the same adaptation algorithm $\tF$ for evaluation. %
All $\tF$ are trained with 5-shot data.
In~\tabref{table:main-clip},
we report $\Delta@2$ on CIFAR100 and ImageNet over the first 9 superclasses. %
On both datasets, introducing $F$ for an intermediate $\widetilde \theta$ generally brings a consistent improvement on $\Delta@2$. This suggests LTO is effective at obstructing the restricted classes without hurting other classes. %
Overall, on ImageNet %
we can obstruct the performance of CE in $\gR$ by 7.65\%, CoOp by 4.58\%, and Tip-Adapter by 5.86\%.\\
\indent\myparagraph{Data efficiency.}
It is intuitive that after applying LTO, if more data is available to the FSC $\gT$, then can overcome the obstruction. 
In~\tabref{table:data-clip}, we test out this scenario by applying more data in the FSC methods for evaluation, denoted as $\tF'$, instead of using the same amount of data for both $\tF$ and $\tF'$.
We study the performance on superclass $\text{id}=1$. We note that $\gD_\tA$ is still a 5-shot set. A data multiplier of value $m_{\texttt{data}}\times$ denotes that the training data of $\tF'$ is $(5m_{\texttt{data}})$-shot. We observe that for CE and CoOp~\cite{zhou2022learning}, it is impactful for them to use more training data, yet it cannot fully overcome the obstruction. In the case for Tip-Adapter~\cite{zhang2022tip}, the difference between using FSC with $20$-shot and 5-shot is $2.24\%$ on $\Delta@2$.
We can conclude that the $\tF'$ for evaluation must leverage a lot more data than that $\tF$ in LTO had used to recover the degraded accuracy. Even then, only a partial of that performance is recovered.\\
\indent\myparagraph{Time Efficiency.}
Next, we study whether training longer in the FCS methods can 
overcome the obstruction caused by LTO. 
In~\tabref{table:time-clip}, we report the $\Delta@2$ when FSG uses more training epochs. An epoch multiplier of value $m_{\texttt{epoch}}\times$ denotes that the training epoch of FSG is $m$ times the one it originally was. We observe that although increasing the training epoch decreases $\Delta@2$, 
the gap is not fully recoverable. Especially for $\texttt{Tip-Adapter}$~\cite{zhang2022tip}, by training for $4\times$ longer for, the improvement is a $2.76\%$ on $\Delta@2$. However, in the case of $\texttt{CoOp}$~\cite{zhou2022learning}, it is impactful for them to use more training epochs. 
It is shown that the $\tF'$ for evaluation must have access to a lot more computation resources than what $\tF$ had used to overcome some of the degraded performance.

\myparagraph{Mismatched FSC.} All previous experiments assume, we use the same FSG method in our LTO and during evaluation. %
We now experiment with a mismatch in FSG, \ie, we use $\tF$ in LTO and another FSC method $\tF'$ for evaluation.
In~\tabref{table:cross}, we report the $\Delta@2$ on different combinations of $(\tF, \tF')$. We observe that the drop ratio is indifferent to which $\tF$ was used during the obstruction. %
When the LTO uses $\texttt{Tip-adapter}$, the difference of $\Delta@2$ between $\tF'=\texttt{CE}$ and $\tF'=\texttt{Tip-adapter}$ is only $1.76\%$ while $\tF'=\texttt{CoOp}$ is actually more vulnerable and gain $1.83\%$ on $\Delta@2$.
\begin{table*}[t!]
    \begin{minipage}[t]{0.32\linewidth}
        \input{tables/clip-data-eff}

    \end{minipage}
    \hfill
    \begin{minipage}[t]{0.32\linewidth}
        \input{tables/clip-time-eff}

    \end{minipage}
    \hfill
    \begin{minipage}[t]{0.32\linewidth}
        \input{tables/cross_atk_def}

    \end{minipage}
    \vspace{-0.6cm}
\end{table*}

%% file: tables/clip_main.tex
\captionof{table}{\textbf{\texttt{DropRatio} of Clip-based LTO on CIFAR100 and ImageNet.} We report $\Delta@2$ on CIFAR100 and ImageNet over 9 selected superclasses.  Note, superclasses are not the same across datasets.%
}
\label{table:main-clip}
\small
\centering
\setlength{\tabcolsep}{5pt}
\resizebox{\linewidth}{!}{
\begin{tabular}{
    c@{\hskip 1pt}
    c@{\hskip 3pt}
    rr
    >{\columncolor{OursColor}}r
    c
    rr
    >{\columncolor{OursColor}}r
}
\specialrule{.15em}{.05em}{.05em}
\multirow{2}{*}{$\tF$} & 
\multirow{2}{*}{$\gR$} &
\multicolumn{3}{c}{CIFAR100} && 
\multicolumn{3}{c}{ImageNet} 
\\
 & & 
\bsla & \bslb & Ours &&
\bsla & \bslb & Ours \\ 
\cmidrule(lr){1-9}
\multirow{10}{*}{\rotatebox[origin=c]{90}{CE}}
& 1 &
3.68 & \bf 17.66 & 14.55 &&
2.79 & 4.82 & \bf 9.93
\\
& 2 &
0.69 & 9.87 & \bf 15.22 &&
2.09 & 0.88 & \bf 4.84
\\
& 3 &
1.71 & 3.88 & \bf 9.96 &&
1.08 & 1.18 & \bf 6.89
\\
& 4 &
0.35 & 2.71 & \bf 15.99 &&
2.01 & 1.11 & \bf 9.15
\\
& 5 &
1.66 & 4.09 & \bf 4.56 &&
1.20 & 0.96 & \bf 2.72
\\
& 6 &
0.31 & 13.26 & \bf 18.00 &&
7.64 & 6.06 & \bf 15.86
\\
& 7 &
0.84 & 3.05 & \bf 15.64 &&
5.89 & 2.77 & \bf 8.84
\\
& 8 &
0.60 & 1.09 & \bf 10.82 &&
0.97 & 0.78 & \bf 6.59
\\
& 9 &
3.45 & \bf 3.52 & 3.17 &&
1.53 & 2.30 & \bf 4.02
\\
\cmidrule(lr){2-9}
& Avg. &
1.48 & 6.57 & \bf 11.99 &&
2.80 & 2.32 & \bf 7.65
\\
\cmidrule(lr){1-9}
\multirow{10}{*}{\rotatebox[origin=c]{90}{CoOp~\cite{zhou2022learning}}}
& 1 &
6.56 & \bf 8.44 & 5.91 &&
1.15 & 2.23 & \bf 6.15
\\
& 2 &
0.56 & \bf 3.49 & 3.03 &&
1.02 & 2.47 & \bf 8.49
\\
& 3 &
0.83 & 6.56 & \bf 8.96 &&
1.51 & \bf 4.54 & 3.29
\\
& 4 &
0.73 & \bf 7.95 & 6.99 &&
0.57 & 2.97 & \bf 5.55
\\
& 5 &
1.21 & \bf 3.89 & 1.58 &&
1.30 & \bf 2.70 & 1.93
\\
& 6 &
0.31 & 5.94 & \bf 6.80 &&
0.84 & 1.91 & \bf 5.02
\\
& 7 &
1.82 & 5.78 & \bf 11.40 &&
1.13 & 2.84 & \bf 3.91
\\
& 8 &
1.35 & 4.85 & \bf 12.02 &&
1.08 & 1.44 & \bf 3.01
\\
& 9 &
2.83 & \bf 8.44 & 3.87 &&
2.14 & 3.47 & \bf 3.95
\\
\cmidrule(lr){2-9}
& Avg. &
1.80 & 6.15 & \bf 6.73 &&
1.19 & 2.73 & \bf 4.58
\\
\cmidrule(lr){1-9}
\multirow{10}{*}{\rotatebox[origin=c]{90}{Tip-Adapter~\cite{zhang2022tip}}}
& 1 &
3.68 & 9.57 & \bf 15.09 &&
1.37 & 2.78 & \bf 5.92
\\
& 2 &
0.85 & 3.41 & \bf 17.12 &&
2.04 & 3.26 & \bf 4.89
\\
& 3 &
1.93 & \bf 10.21 & 6.39 &&
1.76 & 2.83 & \bf 5.59
\\
& 4 &
0.62 & 1.07 & \bf 7.48 &&
1.76 & 4.00 & \bf 7.49
\\
& 5 &
4.63 & 3.04 & \bf 6.78 &&
1.27 & 2.25 & \bf 3.25
\\
& 6 &
3.45 & 3.66 & \bf 8.64 &&
4.74 & 3.99 & \bf 8.57
\\
& 7 &
0.92 & 4.47 & \bf 11.17 &&
3.10 & 2.36 & \bf 7.41
\\
& 8 &
1.28 & 0.82 & \bf 6.39 &&
1.81 & 1.88 & \bf 4.97
\\
& 9 &
1.35 & 3.65 & \bf 12.39 &&
1.62 & 1.41 & \bf 4.64
\\
\cmidrule(lr){2-9}
& Avg. &
2.08 & 4.43 & \bf 10.16 &&
2.16 & 2.75 & \bf 5.86
\\
\specialrule{.15em}{.05em}{.05em}
\end{tabular}
}

%% file: tables/clip-data-eff.tex
\caption{\textbf{Data efficiency.} We report the $\Delta@2$ of our LTO on ImageNet obstructing superclass $\text{id}$=1 when $|\gD_\tF|$ is $m_\texttt{data}$, i.e. the data multiplier, times of $|\gD_\tA|$.  TA: Tip-Adapter~\cite{zhang2022tip}. CO: CoOp~\cite{zhou2022learning}.}
\vspace{-0.2cm}
\label{table:data-clip}
\centering
\resizebox{\columnwidth}{!}{
\begin{tabular}{crrrr}
\specialrule{.15em}{.05em}{.05em}
\multirow{2}{*}{$\tF$} & \multicolumn{4}{c}{$m_\texttt{data}$ }\\
& \centercell{1$\times$} & \centercell{2$\times$} & \centercell{3$\times$} & \centercell{4$\times$} \\
\cmidrule(lr){1-5}
{CE}
 & 9.93 & 6.34 & 2.76 & 2.82 \\ 
{CO}
 & 6.15 & 2.46 & 2.26 & 2.32 \\
{TA} 
 & 5.92 & 3.09 & 3.06 & 3.76 \\
\specialrule{.15em}{.05em}{.05em}
\end{tabular}
}

%% file: tables/clip-time-eff.tex
\caption{\textbf{Time efficiency.}  We report the $\Delta@2$ of our LTO on ImageNet obstructing superclass $\text{id}$=1 when the training epoch of FCS methods is $m_{\texttt{time}}$, i.e. the epoch multiplier, times of its original. %
}
\vspace{-0.2cm}
\label{table:time-clip}
\centering
\resizebox{\columnwidth}{!}
{
\begin{tabular}{cccccccc}
\specialrule{.15em}{.05em}{.05em}
\multirow{2}{*}{$\tF$} & \multicolumn{4}{c}{$m_{\texttt{time}}$}\\
& 1$\times$ & 2$\times$ & 3$\times$ & 4$\times$ \\
\cmidrule(lr){1-6}
{CE} & 9.93 & 3.06 & 4.66 & 3.64
 \\ 
{CO} & 6.15 & 4.66 & 2.52 & 1.99
\\
{TA} & 5.92 & 2.68 & 3.41 & 3.16
\\
\specialrule{.15em}{.05em}{.05em}
\end{tabular}
}

%% file: tables/cross_atk_def.tex
\captionof{table}{
    \textbf{Cross-$(\tF, \tF')$.} We report the $\Delta@2$ of our LTO on ImageNet obstructing superclass $\text{id}$=1 when the FSC methods $\tF$ and $\tF'$ can mismatch. TA: Tip-Adapter~\cite{zhang2022tip}, CO: CoOp~\cite{zhou2022learning}.
}
\vspace{-0.2cm}
\label{table:cross}
\centering
\resizebox{\columnwidth}{!}{
\begin{tabular}{
    crrrr}
    \specialrule{.15em}{.05em}{.05em}
    \multirow{2}{*}{$\tF$} & 
    \multicolumn{4}{c}{$\tF'$}
    \\
    & \centercell{CE} & \centercell{CO} & \centercell{TA} & \centercell{Avg.}
    \\
    \cmidrule(lr){1-5}
    CE & 
    9.93 & 4.71 & 7.33 & 7.32
    \\
    {CO} &
    4.79 & 6.15 & 4.34 & 5.09
    \\
    {TA} &
    4.16 & 7.75 & 5.92 & 5.94
    \\
    \specialrule{.15em}{.05em}{.05em}
\end{tabular}
}
\centering

%% file: src/exp-03-attr.tex
\subsection{Obstruct CLIP-based Attribute Learning}
\label{sec:exp-attr}

\myparagraph{Experiment setup.}
We select CLIP-ResNet50~\cite{radford2021learning} as the pre-trained backbone. %
We conduct experiments on the CelebA~\cite{liu2015faceattributes} which contains 40 annotated attributes annotated for face images. 
We follow the setup by ~\citet{gannamaneni2023investigating}
and select 12 appearance attributes out of 40, including ``EyeClasses'', ``Wearing\_Hat'', ``Bald'', etc.
In each experiment, 1 out of the 12 selected appearance attributes is selected to be the restricted class$\gR$, while the rest 11 attributes belong to $\gR'$. 
We use the first attribute ($\text{id}=0$) as the validation task for tuning hyperparameters. We report the result for the rest of the attributes from $\text{id}=1 \sim 11$.
To evaluate the performance of attribute learning %
we use AUROC (Area Under the Receiver Operating Characteristic curve) as the main metric.

\myparagraph{FSC method.}
We extend the CE learner from~\secref{sec:exp_clip} to conduct attribute learning. %
Following the naive prompt settings~\cite{gannamaneni2023investigating}, for each attribute $a \in \gA$, we build a binary classifier. The classifier consists of two features, a positive and a negative prompt. 
The classifier makes a prediction $\hat a$ by determining which feature has a higher product with the visual feature of input image $\vv_\vx$.

\begin{figure*}[t]
\centering
\includegraphics[width=.975\linewidth]{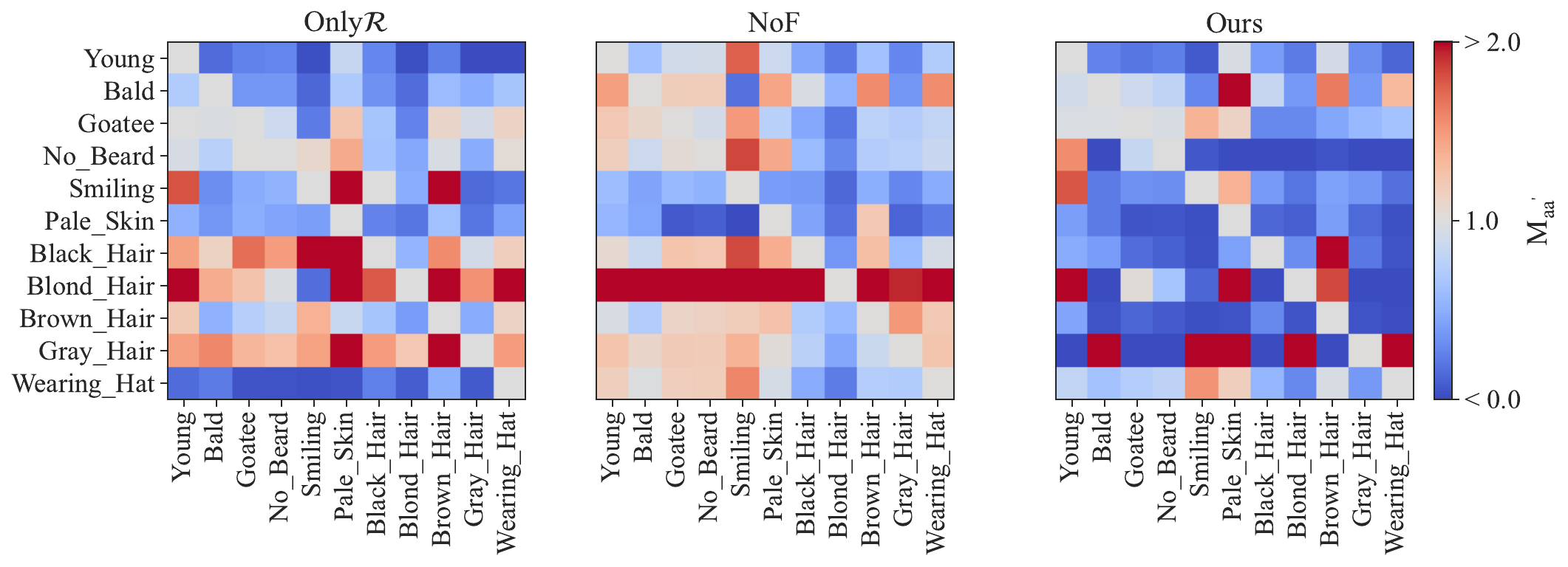}
\vspace{-0.4cm}
\caption{
    \textbf{Confusion matrix on CelebA attribute learning.} 
    Each cell $M_{aa'} = {\delta^{a'}_{a}}/{\delta^{a}_{a}}$ where $\delta^{i}_{j}$ denotes the decrease of AUROC in attribute $i$ when LTO obstruct $\gR=\{ j \}$ for all $i, j \in \gA$. Ideally, all non-diagonal values should be as small (blue) as possible.
}
\label{fig:attr-confusion}
\vspace{-0.7cm}
\end{figure*}

\myparagraph{Results.} In~\tabref{table:main-celeba}, we report the $\Delta@2$ of LTO with $\tF = \texttt{CE}$ on CelebA attribute learning.
We observe consistent drops of AUROC on all attributes, while some attributes are more vulnerable than others to LTO. Specifically, for ours,  
\begin{wrapfigure}[13]{r}{0.5\linewidth}
\vspace{-0.35cm}
\input{tables/clip_celeba}

\end{wrapfigure}
attribute ``Pale\_Skin''(\text{id}$=6$)  is the easiest to obstruct with $\Delta@2 = 31.64\%$ while attribute `Gray\_Hair''(\text{id}$=10$) is the hardest to obstruct with $\Delta@2\% = 0.08$. In comparison to the baselines, ours also achieves the most significant $\Delta@2$ on most attributes. Overall, our LTO can decrease the average of AUROC by $\Delta@2 = 8.04\%$.

\myparagraph{Confusion among attributes.}
We define the confusion matrix $\mM \in \sR^{|\gA|\times |\gA|}$ as follows, each cell $M_{aa'}$ is the ratio ${\delta^{a'}_{a}}/{\delta^{a}_{a}}$ where $\delta^{i}_{j}$ denotes the decrease of AUROC in attribute $i$ when LTO obstruct $\gR=\{ j \}$ for all $i, j \in \gA$. If $M_{aa'} < 1$, then the performance drop on attribute $a'$ is smaller than the performance drop on attribute $a$, which is a success case for our LTO. On the other hand, if $M_{aa'} > 1$, then the performance drop on attribute $a'$ is larger than the performance drop on attribute $a$, which is considered a failure of obstruction. %
This visualization also shows whether two attributes $a$ and $a'$ are related during obstruction, \ie, did LTO obstruct $a$ while unintentionally also obstructing $a'$?

In~\figref{fig:attr-confusion}, we visualize the confusion matrices of \bsla, \bslb, and ours on the 12 selected attributes. Ideally, any non-diagonal $M_{aa'}$, where $a \neq a'$, should be as small (blue) as possible. Visually speaking, our LTO method outperforms the two baselines. \method successfully lowers the performance of restricted attributes during obstruction learning.
We also observed that certain attribute pairs with a strong correlation in the confusion matrix of our LTO. For example, it is hard to obstruct ``Blond\_hair'' without obstructing ``Brown\_hair'', which is understandable since the colors are similar. There are also high correlation between (``Gray\_Hair'', ``Bald'') or (``No\_beard'', ``Young'').

%% file: tables/clip_celeba.tex
\captionof{table}{\textbf{CLIP-based LTO on CelebA attribute learning.} We report the $\Delta@5$ for each selected attributes.
}
\vspace{-0.03cm}
\label{table:main-celeba}
\centering
\setlength{\tabcolsep}{5pt}
\resizebox{\linewidth}{!}{
\begin{tabular}{
    c@{}
    c@{}c
    >{\columncolor{OursColor}}c
    c
    c@{}
    c@{}c
    >{\columncolor{OursColor}}c
}
\specialrule{.15em}{.05em}{.05em}
{$\gR$} & 
\centercell{\bsla} & \centercell{\bslb} & Ours&&
{$\gR$} &
\centercell{\bsla} & \centercell{\bslb} & Ours 
\\
\cmidrule(lr){1-9}
1 &
\bf 5.56 & 2.96 & 4.04 &&
7 & 
1.69 & \bf 2.48 & 2.38
\\
2 &
1.01 & 3.58 & \bf 5.20 &&
8 & 
0.77 & 0.48 & \bf 1.62
\\
3 & 
0.79 & 0.54 & \bf 1.11 &&
9 & 
7.26 & 4.13 & \bf 10.01
\\
4 & 
0.29 & 0.57 & \bf 9.48 &&
10 &
\bf 0.87 & 0.72 & 0.08
\\
5 & 
0.75 & 0.96 & \bf 1.96 &&
11 & 
\bf 7.67 & 1.07 & 4.94
\\
6 & 
5.40 & 10.38 & \bf 31.64 &&
Avg. 
& 3.56 & 3.10 & \bf 8.04
\\
\specialrule{.15em}{.05em}{.05em}
\end{tabular}
}

%% file: src/exp-05-unlearning.tex
\myparagraph{Additional comparison with unlearning for obstruction.}
While machine unlearning is not designed for obstruction, the unlearning algorithms can be used to modify a pre-trained model. In this section, we benchmark against machine unlearning and show that LTO is not just unlearning the classes, but also \textit{making them more difficult to learn back}. We choose SSD~\cite{foster2023fast}, a state-of-the-art machine unlearning algorithm on CIFAR100 for comparison. The experiment setup follows the setting in~\secref{sec:exp_clip}. We apply SSD on the pre-trained CLIP-RN50 weights
\begin{wrapfigure}[36]{tr}{0.45\linewidth}%
\vspace{-1.25cm}%
\input{tables/clip_ssd}\end{wrapfigure}%
by setting the forget set to be the set of restricted classes $\gR$, while the retain set is set to $\gR'$.   
Note that we are considering, \ie, %
if $\gD_A$ is 5-shot then forget and retrain set are also 5-shot.

With the pre-trained model unlearned, we then apply few-shot algorithms $\tF \in \{{\tt CE}, {\tt CoOp}, {\tt TipAdapter}\}$ with $100$-way 5-shot to learn over both $\gR$ and $\gR'$. %
We report the drop ratio $\Delta@2$ as in~\tabref{table:main-clip}. As shown in~\tabref{table:unlearn-clip}, the results on the first 9 superclasses in CIFAR100 show the averages $\Delta@2$ of SSD are 1.09/1.17/1.07 for each $\tF$ while ours are 11.99/6.73/10.16. We observe that SSD is not able to unlearn the forget set while maintaining the performance on the retain set; We suspect that this is because SSD is not designed for the few-shot setting. Hence, we then increase $\gD_A$ for both methods to 20-shot to study how the size of the training set affects their performance. Although SSD benefits from utilizing more training data, 20-shot SSD is still not comparable to 5-shot ours on most superclasses. Moreover, 20-shot LTO outperforms 20-shot SSD significantly under all scenarios. %

%% file: tables/clip_ssd.tex
\captionof{table}{\textbf{\texttt{DropRatio} of unlearning method SSD~\cite{foster2023fast} versus our LTO on CIFAR100.} 
We report $\Delta@2$ on CIFAR100 over 9 selected superclasses. Each method is trained on $\gD_A$ under two few-shot settings, i.e. 5-shot and 20-shot, but evaluated on 5-shot $\tF$.
}
\label{table:unlearn-clip}
\footnotesize
\centering
\setlength{\tabcolsep}{3pt}
\begin{tabular}{
    c@{\hskip 1pt}
    c@{\hskip 3pt}
    r
    >{\columncolor{OursColor}}r
    r
    >{\columncolor{OursColor}}r
}
\specialrule{.15em}{.05em}{.05em}
\multirow{2}{*}{$\tF$} & \multirow{2}{*}{$\gR$} &
\multicolumn{2}{c}{5-shot $\gD_A$} & \multicolumn{2}{c}{20-shot $\gD_A$}
\\
 & & \centercell{SSD} & \centercell{\cellcolor{OursColor}Ours} & \centercell{SSD} & \centercell{\cellcolor{OursColor}Ours}
\\
\cmidrule(lr){1-6}
\multirow{10}{*}{\rotatebox[origin=c]{90}{CE}}
& 1 &
0.67 & \bf 14.55 & 2.22 & \bf 21.87
\\
& 2 &
1.28 & \bf 15.22 & 7.62 & \bf 34.41
\\
& 3 &
1.17 & \bf 9.96 & 0.36 & \bf 15.23
\\
& 4 &
1.31 & \bf 15.99  & 1.69 & \bf 12.25
\\
& 5 &
1.23 & \bf 4.56 & 1.23 & \bf 15.29
\\
& 6 &
1.01 &  \bf 18.00 & 7.57 & \bf 24.34
\\
& 7 &
1.03 & \bf 15.64 & 1.14 & \bf 18.62
\\
& 8 &
0.64 & \bf 10.82 & 2.76 & \bf 18.14
\\
& 9 &
 1.42 & \bf 3.17 & 2.23 & \bf 20.28
\\
\cmidrule(lr){2-6}
& Avg. &
1.09 & \bf 11.99 & 2.98 & \bf 20.04
\\
\cmidrule(lr){1-6}
\multirow{10}{*}{\rotatebox[origin=c]{90}{CoOp~\cite{zhou2022learning}}}
& 1 &
0.74 & \bf 5.91 & 3.72 & \bf 21.25
\\
& 2 &
1.23 & \bf 3.03 & 6.68 & \bf 16.81
\\
& 3 &
1.18 & \bf 8.96  & 3.16 & \bf 9.33
\\
& 4 &
1.30 & \bf 6.99 & 7.60 & \bf 14.28
\\
& 5 &
1.31 & \bf 1.58 & 1.33 & \bf 11.57
\\
& 6 &
1.11 & \bf 6.80 & 10.03 & \bf 20.00
\\
& 7 &
0.88 & \bf 11.40 & 2.62 &  \bf 11.81
\\
& 8 &
0.68 & \bf 12.02 & 3.12 & \bf 26.32
\\
& 9 &
1.35 & \bf 3.87 & 2.89 & \bf 16.77
\\
\cmidrule(lr){2-6}
& Avg. &
1.17 & \bf 6.73 & 4.57 & \bf 16.46
\\
\cmidrule(lr){1-6}
\multirow{10}{*}{\rotatebox[origin=c]{90}{Tip-Adapter~\cite{zhang2022tip}}}
& 1 &
0.69 & \bf 15.09 & 3.30 & \bf 11.62
\\
& 2 &
1.09 & \bf 17.12 & 4.57 & \bf 9.31
\\
& 3 &
1.17 & \bf 6.39 & 3.38 & \bf 9.30
\\
& 4 &
1.27 & \bf 7.48 & 4.00 & \bf 6.67
\\
& 5 &
1.31 & \bf 6.78 & 1.32 & \bf 9.82
\\
& 6 &
1.11 & \bf 8.64 & 7.63 & \bf 14.89
\\
& 7 &
0.85 & \bf 11.17 & 1.83 &  \bf 12.89
\\
& 8 &
0.76 & \bf 6.39 & 3.18 & \bf 14.17
\\
& 9 &
1.34 & \bf 12.39 & 2.73 & \bf 7.58
\\
\cmidrule(lr){2-6}
& Avg. &
1.07 & \bf 10.16 & 3.55 & \bf 10.69
\\
\specialrule{.15em}{.05em}{.05em}
\end{tabular}

%% file: src/conc.tex
\section{Conclusion}
 We propose LTO that learns a \textit{poor initialization} of the backbone to obstruct FSC on restricted classes. %
 Empirically, LTO successfully obstructs four FSC methods over image classification and attribute classification tasks.
Please note that we study the obstruction of FSC with a more significant objective in mind. We aim to ensure the safer release of open-source models. By developing pre-trained models that are difficult to fine-tune for certain downstream tasks, we can potentially prevent the misuse of released models on known harmful applications, while maintaining the open-source culture of the computer vision community. We believe that LTO on FSC represents a promising step towards this goal.

%% file: src/supp.tex
\onecolumn

\appendix
\newcommand{\beginsupplementary}{%
    \setcounter{section}{0}
	\renewcommand{\thesection}{A\arabic{section}}
	\renewcommand{\thesubsection}{\thesection.\arabic{subsection}}

	\renewcommand{\thetable}{A\arabic{table}}%
	\setcounter{table}{0}

	\renewcommand{\thefigure}{A\arabic{figure}}%
	\setcounter{figure}{0}
	
	\setcounter{algorithm}{0}
}
\beginsupplementary

{\noindent\bf \LARGE Appendix
}
\vspace{0.2cm}
\begin{itemize}
    \item In~\secref{sec:supp_results}, we provide additional results of LTO on CIFAR100~\cite{krizhevsky2009learning} and SUN397~\cite{Xiao:2010}, including more comparisons with machine unlearning and data poisoning baselines.
    \item In~\secref{sec:supp_impl}, we provide additional implementation and experimental details. We will release the code upon acceptance of the paper.
\end{itemize}

\section{Additional Results}\label{sec:supp_results}
We first provide additional results on CIFAR100 for the remaining superclasses ($\text{id} \in [10, 19]$).
Next, we conduct additional experiments on the SUN397~\cite{Xiao:2010} dataset to evaluate the performance of LTO.
The restricted classes $\gR$ are based on the hierarchy tree provided by~\citet{Xiao:2010}. We select 9 superclasses in total. The details are documented in~\secref{sec:selection-R-details}.

\input{tables/cls_fsc_sun397}
\subsection{Classical FSC}

In~\tabref{table:cls_fsc_sun397}, we report the \texttt{DropRatio} $\Delta@2$ on SUN397~\cite{Xiao:2010} over 9 superclasses detailed in~\tabref{table:sun397-superclass}. For the column of \method, we observe that on average $\Delta@2$ is larger than 1 in all settings, indicating our method is effective on the SUN397 dataset. When compared with the baseline methods, \method has the largest $\Delta@2$ under all settings on average, which means our method can better obstruct the model performance in $\gR$ without compromising the other classes in $\gR'$.

\input{tables/clip_cifar_full}

\subsection{CLIP-based FSC}
\myparagraph{CIFAR100.}
In~\tabref{table:clip-fsc-CIFAR100-full}, we report $\Delta@2$ on CIFAR100 over all the superclasses. We observe the consistent improvement between ours and the two baselines. On CIFAR100, LTO achieves an average DropRatio of 10.76, 4.80, and 9.80 for CE, CoOP, and Tip-Adapter respectively.

\input{tables/clip-LTO-sun397}
\myparagraph{SUN397.}
In~\tabref{table:main-clip-sun397}, we report $\Delta@2$ on SUN397 over 9 selected superclasses. 
On SUN397, LTO achieves an average DropRatio of 12.27, 4.26, and  5.73 for CE, CoOP, and Tip-Adapter respectively.


\input{tables/clip_ssd_full}

\subsection{Machine Unlearning}
Following the experiment setup in ~\tabref{table:unlearn-clip}, we report the results of unlearning method SSD~\cite{foster2023fast} versus our LTO on CIFAR100 over all the superclasses in ~\tabref{table:unlearn-clip-full}.
Over all the superclasses, we observe that SSD performs poorly under the 5-shot setting in comparison to our LTO. After increasing $\gD_A$ for both methods to 20-shot, 20-shot SSD is comparable to 5-shot ours on a few superclasses. However, the drop ratios of 20-shot are very inconsistent across the 19 superclasses. Moreover, 20-shot LTO outperforms 20-shot SSD significantly under most scenarios. 

\subsection{Data Poisoning}
In this section, we provide additional comparisons to data poisoning which is not designed for the task of learning to obstruct but can potentially be used for this task.
Inspired by the setting of~\citet{oldewage2022adversarial}, we designed data poisoning experiments on FSC. For classical FSC, we consider poisoning the data for restricted classes $\gR$ in both the support and query sets of meta-training as the users will not use poisoned data for meta-testing. We obtained the poisoned data by running ASP~\cite{oldewage2022adversarial} with ProtoNet~\cite{snell2017prototypical}. After that, we paired the poisoned images with labels, predicted by the pre-trained Inception-V3, to be used for meta-training on ProtoNet and MetaOptNet~\cite{lee2019meta}. 
For CLIP-based FSC, we poison the data for restricted classes in $\gD_A$.
By ablating the ratio of poisoned data vs. clean data, we found that in both FSC settings, when the accuracy on other classes $\gR'$ in average drops by 2\% then on restricted $\gR$ also drops by 2\% in average, \ie, the \texttt{DropRatio} $\Delta$@2$\approx$1. This means that data poisoning offers a weaker obstruction quality than LTO.

\section{Implementation Details}\label{sec:supp_impl}

\subsection{Classical FSC}

\myparagraph{Data details.} Across all experiments, we use 70\% of the images in each class for LTO training and use the rest 30\% for few-shot classification $\tF$ and evaluation. In the LTO training process, we generate 1,000 different tasks, and within each task, both $\gD_{\tt fsc}$ and $\gD_{\tt obs}$ contain 15 images for the query set. Among the five classes of each task, there is one class in $\gR$ and the other four classes in $\gR'$.

\myparagraph{Training details.} Across all experiments, we generate 1,000 different tasks and run 20 steps of $F$ with the batch size $|\gB| = 20$. When $\tF$ is ProtoNet, the learning rate of training the learner $F$ is $5.0e^{-5}$ and the learning rate $\alpha$ for LTO training is 0.01 for 1-shot and 0.015 for 5-shot. When $\tF$ is MetaOptNet, the learning rate for $F$ is also $5.0e^{-5}$ and the learning rate $\alpha$ for LTO training is $5.0e^{-5}$ for both 1-shot and 5-shot settings. 

\myparagraph{Computation requirement.}
All experiments are conducted on a single NVIDIA A6000 GPU with 48GB memory. Under the settings described above, for both ProtoNet~\cite{snell2017prototypical} and MetaOptNet~\cite{lee2019meta}, we run 200 steps of LTO training where each step takes 28s on both datasets. While for FSC $\tF$, the training time for one epoch with 512 tasks is 90s.

\subsection{Clip-based FSC}

\myparagraph{Computation Requirement.}
All experiments are conducted on a single NVIDIA A6000 GPU with 48GB memory. 
We report the comparison of training time between our LTO and CLIP-based FSC on each dataset in~\tabref{table:compute-clip-based-fsc}. The training time of our LTO is roughly $10 \times$ to $40\times$ the training time of corresponding FSC depending on the dataset.

\input{tables/compute-clip-based-fsc}

\myparagraph{Choices of FSC $\tF$.}
In~\secref{sec:exp_clip}, we consider a simple learner, namely Cross-Entropy (CE). The CE learner optimizes the CLIP by minimizing cross-entropy loss without introducing any new parameters.
Additionally, we consider two state-of-the-art CLIP-based few-shot learning methods to obstruct: CoOp~\cite{zhou2022learning} and TipAdapter~\cite{zhang2022tip}. 
However, these SOTA few-shot learners fix the pre-trained weights $\theta^p$ and only update the weights of additionally introduced modules $\phi$. This makes these learners extremely vulnerable to LTO since they have little way to overcome the obstruction caused in $\tA(\theta)$.
To make the experiment setup more suitable for these learners, we strengthen them by allowing the gradients to update the CLIP pre-trained weights.
When training LTO for $\tF = $ {\tt TipAdapter}~\cite{zhang2022tip}, we set the number of its augmentation epoch to one. The original setting reported by~\citet{zhang2022tip} is 20. We set its residual ratio $\alpha$ to $1.0$ and sharpness ratio $\beta$ to 5.5 for all experiments.

\myparagraph{Training details.}
For all LTO methods, we only update the non-BN weights in CLIP's image encoder and the token-embedding layer in CLIP's text encoder. For all FSC learners, we update the non-BN weights in CLIP's image encoder, the token-embedding layer in CLIP's text encoder, and whatever additional weights $\phi$ they introduce.
For CE, the learning rate of CLIP's image and text encoder is set to 1e-6 and 5e-5 respectively with a cosine annealing scheduler. For every 4 FSC steps in $\tA$, we resample a new set of $\gP'$ and $\gY'$ where $|\gP'| = 1$ and $|\gY'| = \frac{1}{4} |\gY|$.

For CIFAR100, we run 100 steps of LTO. We set the learning rate $\alpha$ on the visual encoder to be $1e^{-6}$ and the $\alpha$ on the token-embedding layer to be $5e^{-5}$.
For ImageNet, we run 50 steps of LTO. We set the learning rate on the visual encoder to be $1e^{-5}$ and the learning rate on the token-embedding layer to be $5e^{-4}$.
For SUN397, we run 50 steps of LTO. We set the learning rate $\alpha$ on the visual encoder to be $4e^{-6}$ and the learning rate on the token-embedding layer to be $2e^{-4}$.
For all datasets, we set the task batch size $|\gB|$ to be one.

\subsection{Attribute learning}

\myparagraph{Data split details.}
We took 20\% from the CelebA training set and split them in half to form the training set for LTO $\gD_\tA$ and for FSC $\gD_\tF$.

\myparagraph{Training details.}
We run 20 steps of LTO. We set the learning rate $\alpha$ on the visual encoder to be $4e^{-5}$ and the $\alpha$ on the token-embedding layer to be $2e^{-3}$.
We set $|\gB|$ to be 1. For each batch, we sample 1,000 images from $\gD_\tA$ as $\gD_{\texttt{fsc}^{(t)}}$ and another 1,000 images from $\gD_\tA$ as $\gD_{\texttt{obs}^{(t)}}$.

\subsection{Documenting superclass and attribute splits}
\label{sec:selection-R-details}

\myparagraph{CelebA attributes and prompts.}
In~\secref{sec:exp-attr}, when conducting experiments of CLIP-based attribute learning on CelebA~\cite{liu2015faceattributes}, we select 12 appearance attributes out of the total 40 attributes.
For each attribute $a \in \gA$, we build a binary classifier from positive and negative prompts. 
We follow the naive prompts by~\citet{gannamaneni2023investigating}. Each attribute has a positive prompt, usually taking the form of ``a photo of a person with \{$a$\}'', and a negative prompt, usually taking the form of ``a photo of a person without \{$a$\}''.  
In~\tabref{table:celeba-attr}, we detail the selected 12 attributes and the corresponding positive and negative prompts for each attribute.

\input{tables/celeba-attr}

\myparagraph{ImageNet superclasses.}
In~\secref{sec:exp-obs-classifical} and~\secref{sec:exp_clip}, when experimenting on ImageNet~\cite{deng2009imagenet}, we use the superclasses by~\citet{robustness}. We use the balanced form for the superclasses,~\ie, each superclass has the same number of members. We set the number of superclasses to 10 and each superclass consists of 38 classes. In~\tabref{table:imagenet-superclass}, we detail all the members of these 10 superclasses.

\input{tables/imagnet-superclass}

\myparagraph{CIFAR100 superclasses.}
In~\secref{sec:exp_clip}, when experimenting on CIFAR100~\cite{krizhevsky2009learning}, we use the superclasses provided by~\citet{krizhevsky2009learning}. 
In~\tabref{table:cifar100-superclass}, we detail all the members of the 20 superclasses.

\input{tables/cifar100-superclass}

\myparagraph{SUN397 superclasses.}
In ~\secref{sec:supp_results},  when experimenting on SUN397~\cite{Xiao:2010}, we use the superclasses provided by~\citet{Xiao:2010}. All 397 scene categories are labeled with a three-level hierarchy, with the first level of 3 superordinate categories and the second level of 15 basic-level categories.
We select the second level as the base for our superclasses. Some scene class is labeled with two second-level categories,~\eg, {\tt barn} is both a {\tt forest, field, jungle} and a {\tt houses, cabins, gardens, and farms}. In our experiments, we put them in the superclass with the smaller id number,~\eg, {\tt barn} is categorized as the {\tt forest, field, jungle} superclass.

\input{tables/sun397-superclass}

%% file: tables/cls_fsc_sun397.tex
\begin{table}[!h]
\small
\caption{\textbf{\texttt{DropRatio} of Classical LTO on SUN397.} We report $\Delta@2$ on SUN397 over 9 selected superclasses. We note that the first superclass ($\text{id} = 0$) is used for validation and thus not reported here.
}
\label{table:cls_fsc_sun397}
\centering
\setlength{\tabcolsep}{6pt}
\resizebox{\columnwidth}{!}{
\begin{tabular}{
    c
    rr
    >{\columncolor{OursColor}}r
    rr
    >{\columncolor{OursColor}}r
    c
    rr
    >{\columncolor{OursColor}}r
    rr
    >{\columncolor{OursColor}}r
}
\specialrule{.15em}{.05em}{.05em}
\multirow{3}{*}{$\gR$} &
\multicolumn{6}{c}{ProtoNet~\cite{snell2017prototypical}} &&
\multicolumn{6}{c}{MetaOptNet~\cite{lee2019meta}}
\\
&
\multicolumn{3}{c}{1-shot} &
\multicolumn{3}{c}{5-shot} &&
\multicolumn{3}{c}{1-shot} &
\multicolumn{3}{c}{5-shot}
\\
 & 
\bsla & \bslb & Ours &
\bsla & \bslb & Ours &&
\bsla & \bslb & Ours &
\bsla & \bslb & Ours 
\\ 
\cmidrule(lr){1-14}
1 
& 0.88 & 2.42 & \bf 4.17 
& 1.23 &\bf 2.48 & 1.59 &
& 1.10 &\bf 3.63 & 2.36 
& 1.30 &\bf 10.25 & 4.22 
\\
2
& 1.05 & 2.86 &\bf 3.60 
& 1.21 & 1.42 &\bf 2.65 &
& 1.78 & 1.72 &\bf 10.17 
& 0.98 & 7.92 &\bf 26.84 
\\
3 
& 0.87 & 1.16 &\bf 2.06 
& 0.84 & 1.00 &\bf 1.21 &
& 0.97 & 6.03 &\bf 7.49
& 0.94 & 8.19 &\bf 19.00 
\\
4 
& 0.93 & 1.87 &\bf 2.68 
& 1.01 &\bf 1.70 & 1.32 &
& 1.05 & 1.21 &\bf 2.31 
& 0.71 & 2.91 &\bf 5.67 
\\
5
& 1.18 &\bf 3.17 & 2.80 
& 1.35 &\bf 2.02 & 1.63 &
& 1.73 & 4.21 &\bf 9.93 
& 2.20 & 7.03 &\bf 16.01 
\\
6 
& 0.99 &\bf 3.47 & 2.92 
& 1.27 &\bf 1.42 & 1.30 &
& 1.26 & 1.14 &\bf 12.01 
& 0.98 & 1.04 &\bf 33.84 
\\
7 
& 0.73 &\bf 1.03 & 0.39 
& 0.59 & 0.92 &\bf 1.60 &
& 1.59 & 3.59 &\bf 6.49 
& 1.68 & 3.00 &\bf 29.31 
\\
8
& 0.90 &\bf 4.16 & 2.98 
& 1.10 & 1.23 &\bf 1.75 &
& 0.91 &\bf 2.00 & 1.67 
& 0.67 & 11.22 &\bf 13.82 
\\
9 
& 1.07 & 1.46 &\bf 1.83 
& 1.12 & 1.19 &\bf 1.47 &
& 1.26 &\bf 10.77 & 8.27 
& 1.08 & 4.70 &\bf 13.39 
\\
\cmidrule(lr){1-14}
Avg. 
& 0.96 & 2.40 &\bf 2.60 
& 1.08 & 1.49 &\bf 1.61 &
& 1.29 & 3.81 &\bf 6.74 
& 1.17 & 5.58 &\bf 18.01 
\\
\specialrule{.15em}{.05em}{.05em}
\end{tabular}
}

\end{table}

%% file: tables/clip_cifar_full.tex
\begin{table}[h!]
\small
\centering
\setlength{\tabcolsep}{4pt}
\caption{
\textbf{\texttt{DropRatio} of Clip-based LTO on CIFAR100.} We report $\Delta@2$ on CIFAR100 over all the 19 superclasses. 
We note that the first superclass ($\text{id} = 0$) is used for validation and thus not reported here.
}
\label{table:clip-fsc-CIFAR100-full}
\resizebox{\columnwidth}{!}{
\begin{tabular}{
    c
    c@{}
    rr>{\columncolor{OursColor}}r
    c
    c@{}
    rr>{\columncolor{OursColor}}r
    c
    c@{}
    rr>{\columncolor{OursColor}}r
    c
    c@{}
    rr>{\columncolor{OursColor}}r
}
\specialrule{.15em}{.05em}{.05em}
&
{$\gR$} & 
\centercell{\bsla} & \centercell{\bslb} & Ours &&
{$\gR$} &
\centercell{\bsla} & \centercell{\bslb} & Ours &&
{$\gR$} &
\centercell{\bsla} & \centercell{\bslb} & Ours &&
{$\gR$} &
\centercell{\bsla} & \centercell{\bslb} & Ours
\\
\cmidrule(lr){1-20}
\multirow{5}{*}{\rotatebox[origin=c]{90}{CE}}
& 1 & 3.68 & \bf 17.66 & 14.55 &
& 6 & 0.31 & 13.26 & \bf 18.00 &
& 11 & 0.77 & 5.18 & \bf 5.38 &
& 16 & 0.47 & 2.07 & \bf 12.17
\\
& 2 & 0.69 & 9.87 & \bf 15.22 &
& 7 & 0.84 & 3.05 & \bf 15.64 &
& 12 & 1.77 & 2.54 & \bf 11.46 &
& 17 & 2.15 & \bf 7.53 & 4.26
\\
& 3 & 1.71 & 3.88 & \bf 9.96 &
& 8 & 0.60 & 1.09 & \bf 10.82 &
& 13 & 3.69 & \bf 7.25 & 5.70 &
& 18 & 1.10 & \bf 3.94 & 2.63
\\
& 4 & 0.35 & 2.71 & \bf 15.99 &
& 9 & 3.45 & \bf 3.52 & 3.17 &
& 14 & 0.55 & 5.90 & \bf 15.42 &
& 19 & 0.38 & 3.72 & \bf 14.18
\\
& 5 & 1.66 & 4.09 & \bf 4.56 &
& 10 & 9.63 & 13.99 & \bf 18.91 &
& 15 & 1.39 & 2.11 & \bf 6.38 &
& Avg. & 1.85 & 5.97 & \bf 10.76
\\
\cmidrule(lr){1-20}
\multirow{5}{*}{\rotatebox[origin=c]{90}{ CO~\cite{zhou2022learning} }}
& 1 & 6.56 & \bf 8.44 & 5.91 &
& 6 & 0.31 & 5.94 & \bf 6.80 &
& 11 & 1.23 & 0.00 & \bf 6.47 &
& 16 & \bf 3.16 & 0.92 & 2.16
\\
& 2 & 0.56 & \bf 3.49 & 3.03 &
& 7 & 1.82 & 3.56 & \bf 11.40 &
& 12 & 2.39 & \bf 4.14 & 3.21 &
& 17 & 0.58 & 1.13 & \bf 1.82
\\
& 3 & 0.83 & 6.56 & \bf 8.96 &
& 8 & 1.35 & 6.92 & \bf 12.02 &
& 13 & 0.75 & \bf 1.84 & 1.17 &
& 18 & 1.46 & \bf 7.33 & 0.85
\\
& 4 & 0.73 & \bf 7.95 & 6.99 &
& 9 & 2.83 & \bf 8.44 & 3.87 &
& 14 & 0.82 & 2.94 & \bf 3.09 &
& 19 & 0.00 & \bf 3.74 & 3.54
\\
& 5 & 1.21 & \bf 3.89 & 1.58 &
& 10 & 3.00 & 2.23 & \bf 7.95 &
& 15 & \bf 3.01 & 1.11 & 0.56 &
& Avg. & 1.72 & 4.54 & \bf 4.80
\\
\cmidrule(lr){1-20}
\multirow{5}{*}{\rotatebox[origin=c]{90}{TA~\cite{zhang2022tip}}}
& 1 & 3.68 & 9.57 & \bf 15.09 &
& 6 & 3.45 & 3.66 & \bf 8.64 &
& 11 & 2.24 & \bf 12.74 & 9.45 &
& 16 & 0.06 & 2.36 & \bf 4.89
\\
& 2 & 0.85 & 3.41 & \bf 17.12 &
& 7 & 0.92 & 4.47 & \bf 11.17 &
& 12 & 0.89 & 4.72 & \bf 17.59 &
& 17 & 1.11 & 2.28 & \bf 5.33
\\
& 3 & 1.93 & 10.21 & 6.39 &
& 8 & 1.28 & 0.82 & \bf 6.39 &
& 13 & 1.45 & 1.10 & \bf 5.54 &
& 18 & 0.33 & 4.44 & \bf 4.54
\\
& 4 & 0.62 & 1.07 & \bf 7.48 &
& 9 & 1.35 & 3.65 & \bf 12.39 &
& 14 & 0.74 & 9.66 & \bf 10.41 &
& 19 & 1.54 & 3.13 & \bf 19.58
\\
& 5 & 4.63 & 3.04 & \bf 6.78 &
& 10 & 8.70 & 5.44 & \bf 5.62 &
& 15 & 0.13 & 0.55 & \bf 11.85 &
& Avg. & 1.89 & 4.54 & \bf 9.80
\\
\specialrule{.15em}{.05em}{.05em}
\end{tabular}
}

\end{table}

%% file: tables/clip-LTO-sun397.tex
\begin{table}[t]
\centering
\setlength{\tabcolsep}{6pt}
\caption{\textbf{\texttt{DropRatio} of Clip-based LTO on SUN397.} We report $\Delta@2$ on SUN397 over 9 selected superclasses. We note that the first superclass ($\text{id} = 0$) is used for validation and thus not reported here.
}
\label{table:main-clip-sun397}
\resizebox{\columnwidth}{!}{
\begin{tabular}{
    c
    rr
    >{\columncolor{OursColor}}r
    c
    rr
    >{\columncolor{OursColor}}r
    c
    rr
    >{\columncolor{OursColor}}r
}
\specialrule{.15em}{.05em}{.05em}
\multirow{2}{*}{$\gR$} &
\multicolumn{3}{c}{CE} &&
\multicolumn{3}{c}{CoOp~\cite{zhou2022learning}} && 
\multicolumn{3}{c}{Tip-Adapter~\cite{zhang2022tip}} 
\\
 & 
\bsla & \bslb & Ours &&
\bsla & \bslb & Ours &&
\bsla & \bslb & Ours 
\\ 
\cmidrule(lr){1-12}
1 & 
3.69 & 3.16 & \bf 11.28 && 
\bf 3.38 & 2.78 & 2.51 && 
1.68 & 2.82 & \bf 3.35
\\
2 & 
4.08 & 3.88 & \bf 7.06 && 
\bf 3.38 & 3.06 & 0.21 && 
3.74 & 3.16 & \bf 4.85
\\
3 & 
3.53 & 3.79 & \bf 5.34 && 
1.49 & 0.89 & \bf 2.10 && 
2.90 & \bf 3.23 & 2.66
\\
4 & 
3.58 & 2.56 & \bf 13.93 && 
3.82 & 2.37 & \bf 5.45 && 
2.50 & 3.68 & \bf 8.42
\\
5 & 
4.66 & 4.24 & \bf 11.31 &&
3.99 & 2.74 & \bf 4.53 && 
4.91 & 4.27 & \bf 6.89
\\
6 & 
9.37 & 9.79 & \bf 15.40 && 
3.40 & 3.12 & \bf 9.20 && 
6.39 & \bf 10.00 & 2.93
\\
7 & 
11.84 & 12.45 & \bf 17.93 && 
3.36 & \bf 4.56 & 4.54 && 
7.69 & 6.07 & \bf 8.03
\\
8 & 
2.55 & 2.70 & \bf 15.91 && 
0.96 & 1.39 & \bf 4.16 && 
4.07 & 5.29 & \bf 6.32
\\
9 & 
7.55 & 5.92 & \bf 12.32 && 
2.52 & 1.67 & \bf 5.65 && 
5.84 & 4.44 & \bf 8.14
\\
\cmidrule{1-12}
Avg. & 
5.65 & 5.39 & \bf 12.27  && 
2.92 & 2.51 & \bf 4.26 && 
4.41 & 4.77 & \bf 5.73
\\
\specialrule{.15em}{.05em}{.05em}
\end{tabular}
}

\end{table}

%% file: tables/clip_ssd_full.tex
\begin{table}[h!]
\small
\centering
\setlength{\tabcolsep}{4pt}
\caption{
\textbf{\texttt{DropRatio} of unlearning method SSD~\cite{foster2023fast} versus our Clip-based LTO on CIFAR100.} We report $\Delta@2$ on CIFAR100 over all the 19 superclasses. 
We note that the first superclass ($\text{id} = 0$) is used for validation and thus not reported here. Each method is trained on $\gD_A$ under two few-shot settings, i.e. 5-shot and 20-shot, but evaluated on 5-shot $\tF$.
}
\label{table:unlearn-clip-full}
\resizebox{\columnwidth}{!}{
\begin{tabular}{
    c
    >{\columncolor{lightgray}}c@{}
    r>{\columncolor{OursColor}}r
    r>{\columncolor{OursColor}}r
    >{\columncolor{lightgray}}c@{}
    r>{\columncolor{OursColor}}r
    r>{\columncolor{OursColor}}r
    >{\columncolor{lightgray}}c@{}
    r>{\columncolor{OursColor}}r
    r>{\columncolor{OursColor}}r
    >{\columncolor{lightgray}}c@{}
    r>{\columncolor{OursColor}}r
    r>{\columncolor{OursColor}}r
}
\specialrule{.15em}{.05em}{.05em}
\multirow{2}{*}{$\tF$} & \cellcolor{white} & \multicolumn{2}{c}{5-shot $\gD_A$} & \multicolumn{2}{c}{20-shot $\gD_A$} &
\cellcolor{white} & \multicolumn{2}{c}{5-shot $\gD_A$} & \multicolumn{2}{c}{20-shot $\gD_A$} &
\cellcolor{white} & \multicolumn{2}{c}{5-shot $\gD_A$} & \multicolumn{2}{c}{20-shot $\gD_A$} &
\cellcolor{white} & \multicolumn{2}{c}{5-shot $\gD_A$} & \multicolumn{2}{c}{20-shot $\gD_A$}
\\
 & \cellcolor{white} \multirow{-2}{*}{$\gR$}  &
\centercell{SSD} & \centercell{\cellcolor{OursColor}Ours} & \centercell{SSD} & \centercell{\cellcolor{OursColor}Ours} &
\cellcolor{white} \multirow{-2}{*}{$\gR$} & 
\centercell{SSD} & \centercell{\cellcolor{OursColor}Ours} & \centercell{SSD} & \centercell{\cellcolor{OursColor}Ours} &
\cellcolor{white} \multirow{-2}{*}{$\gR$} &
\centercell{SSD} & \centercell{\cellcolor{OursColor}Ours} & \centercell{SSD} & \centercell{\cellcolor{OursColor}Ours} &
\cellcolor{white} \multirow{-2}{*}{$\gR$} &
\centercell{SSD} & \centercell{\cellcolor{OursColor}Ours} & \centercell{SSD} & \centercell{\cellcolor{OursColor}Ours} 
\\
\cmidrule(lr){1-21}
\multirow{5}{*}{\rotatebox[origin=c]{90}{CE}}
& 1 & 0.67 & \bf 14.55 & 2.22 & \bf 21.87
& 6 & 1.01 &  \bf 18.00 & 7.57 & \bf 24.34
& 11 & 0.95 & \bf 5.38 & 21.88 & \bf 29.24
& 16 & 0.70 & \bf 12.17 & 2.50 & \bf 16.81
\\
& 2 & 1.28 & \bf 15.22 & 7.62 & \bf 34.41
& 7 & 1.03 & \bf 15.64 & 1.14 & \bf 18.62
& 12 & 0.76 & \bf 11.46 & 1.33 & \bf 22.81
& 17 & 0.99 & \bf 4.26 & 2.21 & \bf 49.23
\\
& 3 & 1.17 & \bf 9.96 & 0.36 & \bf 15.23
& 8 & 0.64 & \bf 10.82 & 2.76 & \bf 18.14
& 13 & 0.85 & \bf 5.70 & 0.92 & \bf 22.44
& 18 & 1.39 & \bf 2.63 & 1.40 & \bf 16.00
\\
& 4 & 1.31 & \bf 15.99  & 1.69 & \bf 12.25
& 9 & 1.42 & \bf 3.17 & 2.23 & \bf 20.28
& 14 & 0.90 & \bf 15.42 & 4.17 & \bf 30.87
& 19 & 1.31 & \bf 14.18 & 1.70 & \bf 13.61
\\
& 5 & 1.23 & \bf 4.56 & 1.23 & \bf 15.29
& 10 & 1.25 & \bf 18.91 & 6.51 & \bf 48.80
& 15 & 0.56 & \bf 6.38 & 2.54 & \bf 56.80
& Avg. & 1.02 & \bf 10.76 & 3.78 & \bf 24.79
\\
\cmidrule(lr){1-21}
\multirow{5}{*}{\rotatebox[origin=c]{90}{ CO~\cite{zhou2022learning} }}
& 1 & 0.74 & \bf 5.91 & 3.72 & \bf 21.25
& 6 & 1.11 & \bf 6.80 & 10.03 & \bf 20.00
& 11 & 1.11 & \bf 6.47 & \bf 23.80 &  4.31
& 16 & 0.71 & \bf 2.16 & 8.99 & \bf 16.38
\\
& 2 & 1.23 & \bf 3.03 & 6.68 & \bf 16.81
& 7 & 0.88 & \bf 11.40 & 2.62 & \bf 11.81
& 12 & 0.82 & \bf 3.21 & 2.19 & \bf 11.76
& 17 & 0.99 & \bf 1.82 & 2.22 & \bf 16.47
\\
& 3 & 1.18 & \bf 8.96  & 3.16 & \bf 9.33
& 8 & 0.68 & \bf 12.02 & 3.12 & \bf 26.32
& 13 & 0.87 & \bf 1.17 & 0.83 & \bf 12.15
& 18 & 1.36 & \bf 0.85 & 1.49 & \bf 4.71
\\
& 4 & 1.30 & \bf 6.99 & 7.60 & \bf 14.28
& 9 & 1.35 & \bf 3.87 & 2.89 & \bf 16.77
& 14 & 1.29 & \bf 3.09 & \bf 5.94 & 3.92
& 19 & 1.17 & \bf 3.54 & 2.26 & \bf 6.67
\\
& 5 & 1.31 & \bf 1.58 & 1.33 & \bf 11.57
& 10 & 1.25 & \bf 7.95 & 6.73 & \bf 12.55
& 15 & \bf 0.67 & 0.56 & \bf 6.86 & 5.23
& Avg. & 1.09 & \bf 4.80 & 5.39 & \bf 12.75
\\
\cmidrule(lr){1-21}
\multirow{5}{*}{\rotatebox[origin=c]{90}{TA~\cite{zhang2022tip}}}
& 1 & 0.69 & \bf 15.09 & 3.30 & \bf 11.62
& 6 & 1.11 & \bf 8.64 & 7.63 & \bf 14.89
& 11 & 1.19 & \bf 9.45 & 36.41 & \bf 43.33
& 16 & 0.75 & \bf 4.89 & 6.71 & \bf 11.37
\\
& 2 & 1.09 & \bf 17.12 & 4.57 & \bf 9.31
& 7 & 0.85 & \bf 11.17 & 1.83 &  \bf 12.89
& 12 & 0.58 & \bf 17.59 & 2.08 & \bf 5.66
& 17 & 0.97 & \bf 5.33 & 2.27 & \bf 10.42
\\
& 3 & 1.17 & \bf 6.39  & 3.38 & \bf 9.30
& 8 & 0.76 & \bf 6.39  & 3.18 & \bf 14.17
& 13 & 0.89 & \bf 5.54 & 0.80 & \bf 12.51
& 18 & 1.35 & \bf 4.54 & 1.43 & \bf 19.23
\\
& 4 & 1.27 & \bf 7.48 & 4.00 & \bf 6.67
& 9 & 1.34 & \bf 12.39 & 2.73 & \bf 7.58
& 14 & 1.16 & \bf 10.41 & 10.05 & \bf 11.37
& 19 & 1.31 & \bf 19.58 & 1.92 & \bf 28.43
\\
& 5 & 1.31 & \bf 6.78 & 1.32 & \bf 9.82
& 10 & 1.12 & \bf 5.62 & 6.87 & \bf 25.67
& 15 & 0.77 & \bf 11.85 & \bf 7.33 & 4.31
& Avg. & 1.03 & \bf 9.80 & 5.67 & \bf 14.13
\\
\specialrule{.15em}{.05em}{.05em}
\end{tabular}
}

\end{table}

%% file: tables/compute-clip-based-fsc.tex
\begin{table}[h!]
\centering
\captionof{table}{
    \textbf{Computation time tradeoff.} We report the rough training time (sec) comparison between our LTO and CLIP-based FSC on CIFAR100, ImageNet, and SUN397. All CLIP-based FSC methods are 5-shot. We also report the average $\Delta@2$ for each dataset to demonstrate the tradeoff.
    TA: Tip-Adapter~\cite{zhang2022tip}, CO: CoOp~\cite{zhou2022learning}.
}
\label{table:compute-clip-based-fsc}
\begin{tabular}{
    c 
    ccc
    c
    ccc
    c
    ccc}
    \specialrule{.15em}{.05em}{.05em}
    \multirow{2}{*}{$\tF$} & 
    \multicolumn{3}{c}{CIFAR100} &&
    \multicolumn{3}{c}{ImageNet} &&
    \multicolumn{3}{c}{SUN397} 
    \\
    & $T_{\tt FSC}$ & $T_{\tt LTO}$ & $\Delta@2$ &
    & $T_{\tt FSC}$ & $T_{\tt LTO}$ & $\Delta@2$ &
    & $T_{\tt FSC}$ & $T_{\tt LTO}$ & $\Delta@2$ 
    \\
    \cmidrule(lr){1-12}
    CE 
    & 40 & 1,000 & 8.93 &
    & 1,200 & 16,000 & 4.76 &
    & 300 & 12,000 & 7.59
    \\
    {CO} 
    & 50 & 1,000 & 8.38 &
    & 2,000 & 18,000 & 6.92 &
    & 350 & 10,000 & 4.46
    \\
    {TA} 
    & 70 & 1,200 & 5.82 &
    & 2,500 & 20,000 & 3.59 &
    & 600 & 15,000 & 4.95
    
    \\
    \specialrule{.15em}{.05em}{.05em}
\end{tabular}
\centering
\end{table}

%% file: tables/celeba-attr.tex
\begin{table}[h!]
\small
\centering
\setlength{\tabcolsep}{3pt}
\caption{\textbf{Selected attributes and prompts for CelebA.} 
We detail the selected attributes for CelebA\cite{liu2015faceattributes} experiments, as well as the positive and negative prompts~\cite{gannamaneni2023investigating} for each attribute. The \ding{55} indicates the negative prompts while the \ding{51} indicates the positive ones.
}
\label{table:celeba-attr}
\resizebox{\columnwidth}{!}{
\begin{tabular}{
    clll
    c
    clll
}
\specialrule{.15em}{.05em}{.05em}
{$\gR$} & {\tt name} & \multicolumn{2}{l}{prompts} & &
{$\gR$} & {\tt name} & \multicolumn{2}{l}{prompts}
\\
\cmidrule{1-9}
\multirow{2}{*}{0} & 
\multirow{2}{*}{Eyeclasses} & 
\ding{55} & a photo of a person not wearing eyeglasses &&
\multirow{2}{*}{6} & 
\multirow{2}{*}{Pale\_Skin} & 
\ding{55} & a photo of a person without pale skin
\\
&&
\ding{51} & a photo of a person wearing eyeglasses &&
&&
\ding{51} & a photo of a person with pale skin
\\
\cmidrule{1-9}
\multirow{2}{*}{1} & 
\multirow{2}{*}{Young} & 
\ding{55} & a photo of a person who is not young &&
\multirow{2}{*}{7} & 
\multirow{2}{*}{Black\_Hair} & 
\ding{55} & a photo of a person with hair that is not black
\\
&&
\ding{51} & a photo of a person who is young &&
&&
\ding{51} & a photo of a person with black hair
\\
\cmidrule{1-9}
\multirow{2}{*}{2} & 
\multirow{2}{*}{Bald} & 
\ding{55} & a photo of a person without a bald head &&
\multirow{2}{*}{8} & 
\multirow{2}{*}{Blond\_Hair} & 
\ding{55} & a photo of a person with hair that is not blond
\\
&&
\ding{51} & a photo of a person with a bald head &&
&&
\ding{51} & a photo of a person with blond hair
\\
\cmidrule{1-9}
\multirow{2}{*}{3} & 
\multirow{2}{*}{Goatee} & 
\ding{55} & a photo of a person without a goatee &&
\multirow{2}{*}{9} & 
\multirow{2}{*}{Brown\_Hair} & 
\ding{55} & a photo of a person with hair that is not brown
\\
&&
\ding{51} & a photo of a person with a goatee &&
&&
\ding{51} & a photo of a person with brown hair
\\
\cmidrule{1-9}
\multirow{2}{*}{4} & 
\multirow{2}{*}{No\_Beard} & 
\ding{55} & a photo of a person with a beard &&
\multirow{2}{*}{10} & 
\multirow{2}{*}{Gray\_Hair} & 
\ding{55} & a photo of a person with hair that is not gray
\\
&&
\ding{51} & a photo of a person without a beard &&
&&
\ding{51} & a photo of a person with gray hair
\\
\cmidrule{1-9}
\multirow{2}{*}{5} & 
\multirow{2}{*}{Smiling} & 
\ding{55} & a photo of a person who is not smiling &&
\multirow{2}{*}{11} & 
\multirow{2}{*}{Wearing\_Hat} & 
\ding{55} & a photo of a person not wearing a hat
\\
&&
\ding{51} & a photo of a person who is smiling &&
&&
\ding{51} & a photo of a person wearing a hat
\\
\specialrule{.15em}{.05em}{.05em}
\end{tabular}
}
\end{table}

%% file: tables/imagnet-superclass.tex
\begin{table}[h!]
\small
\caption{\textbf{Superclasses for ImageNet.} We document the members of the superclasses~\cite{robustness} we used for ImageNet~\cite{deng2009imagenet} experiments. The classes that are not listed in this table are considered to be $\in \gR'$ in all ImageNet experiments.
We set the number of superclasses to 10 and each superclass has 38 classes.
}
\label{table:imagenet-superclass}
\centering
\setlength{\tabcolsep}{4pt}
\resizebox{\columnwidth}{!}{
\begin{tabular}{
    c
    l
    l
}
\specialrule{.15em}{.05em}{.05em}
{$\gR$} & {\tt name} & $\gY_{\tt name}$
\\
\cmidrule{1-3}
0 & 
device & 
\begin{tabular}[t]{l}
French horn, abacus, accordion, acoustic guitar, analog clock, assault rifle, automated teller machine, banjo, barometer, \\ bassoon, bell or wind chime, binoculars, buckle, candle, cannon, car mirror, car wheel, carousel, cello, chainsaw, \\ combination lock, construction crane, cornet, desktop computer, digital clock, digital watch, disc brake, drum, \\ electric fan, electric guitar, flute, gas pump, gong, grand piano, guillotine, hair clip, hand-held computer, hunting bow
\end{tabular}
\\
\cmidrule{1-3}
1 & 
mammal &
\begin{tabular}[t]{l}
African wild dog, Alaskan tundra wolf, American black bear, Angora rabbit, Arctic fox, beaver, brown bear, cheetah, \\ cottontail rabbit, cougar, coyote, dhole, dingo, dugong, fox squirrel, grey fox, grey whale, grey wolf, hamster, hare, \\ hyena, jaguar, killer whale, kit fox, leopard, lion, lynx, marmot, meerkat, mongoose, polar bear, porcupine, red fox, \\ red wolf or maned wolf, sea lion, sloth bear, snow leopard, tiger
\end{tabular}
\\
\cmidrule{1-3}
2 &
dog &
\begin{tabular}[t]{l}
Afghan Hound, Airedale Terrier, American Staffordshire Terrier, Australian Terrier, Basset Hound, Beagle, \\ Bedlington Terrier, Black and Tan Coonhound, Bloodhound, Bluetick Coonhound, Border Terrier, Boston Terrier, \\ Cairn Terrier, Dandie Dinmont Terrier, English foxhound, Giant Schnauzer, Ibizan Hound, Irish Terrier, \\ Irish Wolfhound, Kerry Blue Terrier, Lakeland Terrier, Miniature Schnauzer, Norfolk Terrier, Norwegian Elkhound, \\ Norwich Terrier, Otterhound, Redbone Coonhound,  Rhodesian Ridgeback, Saluki, Scottish Deerhound, \\ Sealyham Terrier, Staffordshire Bull Terrier, Standard Schnauzer, Treeing Walker Coonhound, Weimaraner, \\ Wire Fox Terrier, Yorkshire Terrier, borzoi
\end{tabular}
\\
\cmidrule{1-3}
3 &
structure &
\begin{tabular}[t]{l}
altar, apiary, baluster / handrail, barbershop, barn, boathouse, bookstore, brass memorial plaque, breakwater, \\ butcher shop, candy store, castle, chain-link fence, church, cliff dwelling, dam, dock, fountain, greenhouse, \\ grocery store, home theater, honeycomb, library, lighthouse, megalith, mobile home, monastery,  mosque, movie theater, \\ obelisk, palace, patio, pedestal, picket fence, radiator grille, sawmill, spiral or coil, tent
\end{tabular}
\\
\cmidrule{1-3}
4 &
bird &
\begin{tabular}[t]{l}
American dipper, American robin, african grey parrot, bald eagle, bee eater, black stork, black swan, brambling, \\ bulbul, chickadee, coucal, duck, flamingo, goldfinch, goose, great egret, great grey owl, hen, hornbill, house finch, \\ hummingbird, indigo bunting, jacamar, jay, junco, kite (bird of prey), little blue heron, lorikeet, macaw, magpie, ostrich,\\ red-breasted merganser, rooster, spoonbill, sulphur-crested cockatoo, toucan, vulture, white stork
\end{tabular}
\\
\cmidrule{1-3}
5 &
clothing &
\begin{tabular}[t]{l}
Christmas stocking, T-shirt, abaya, academic gown, apron, bikini, bolo tie, bow tie, bra, cardigan, cowboy hat, \\ crash helmet, diaper, feather boa, football helmet, fur coat, gown, graduation cap, hoop skirt,  jeans, kimono, knee pad, \\ lab coat, mask, military hat (bearskin or shako), military uniform, miniskirt, mitten, one-piece bathing suit, overskirt, \\ pajamas, poke bonnet, poncho, sarong, seat belt, shower cap, swimming cap, tights
\end{tabular}
\\
\cmidrule{1-3}
6 &
arthropod &
\begin{tabular}[t]{l}
American lobster, Dungeness crab, European garden spider, ant, barn spider, bee, centipede, cicada, cockroach, \\ crayfish, cricket insect, dung beetle, fiddler crab, fly, grasshopper, ground beetle, harvestman, hermit crab, isopod, \\ ladybug, leaf beetle, leafhopper, longhorn beetle, praying mantis, red king crab, rhinoceros beetle, rock crab, scorpion,\\ southern black widow, spiny lobster, stick insect, tarantula, tick, tiger beetle, trilobite, weevil, wolf spider, \\ yellow garden spider
\end{tabular}
\\
\cmidrule{1-3}
7 &
matter &
\begin{tabular}[t]{l}
acorn squash, artichoke, bagel, baguette, banana, bell pepper, broccoli, butternut squash, cabbage, carbonara, cardoon, \\ cauliflower, cheeseburger, cherimoya (custard apple), chocolate syrup, consomme, cucumber, fig, guacamole, hay, \\ hot dog, hot pot, ice cream, jackfruit, lemon, mashed potatoes, menu, mushroom, orange, pineapple, plate, pomegranate, \\ popsicle, pretzel, spaghetti squash, strawberry, trifle, zucchini
\end{tabular}
\\
\cmidrule{1-3}
8 &
vehicle &
\begin{tabular}[t]{l}
ambulance, amphibious vehicle, bullock cart, convertible, electric locomotive, fire truck, ford model t, forklift, freight car,\\ garbage truck, go-kart, half-track, horse-drawn vehicle, jeep, limousine, minivan, moped, mountain bike, moving van, \\ pickup truck, police van, race car, railroad car, recreational vehicle, rickshaw, shopping cart, snowmobile, snowplow, \\ sports car, station wagon, steam locomotive, tandem bicycle, tank, taxicab, tow truck, tram, vespa, wheelbarrow
\end{tabular}
\\
\cmidrule{1-3}
9 &
covering &
\begin{tabular}[t]{l}
Band-Aid, Pickelhaube, baby pacifier, balaclava ski mask, bell tower, birdhouse, bottle cap, breastplate, bulletproof vest, \\ chain mail, cloak, clogs, cowboy boot, cuirass, dome, doormat, dust jacket, fire screen, gas mask or respirator, holster, \\ lampshade, lens cap, manhole cover, mosquito net, quilt, ring binder, sandal, scabbard, shield, shoji screen / room divider, \\ slip-on shoe, sneaker, thatched roof, tile roof, umbrella, vaulted or arched ceiling, window screen, window shade
\end{tabular}
\\
\specialrule{.15em}{.05em}{.05em}
\end{tabular}
}
\end{table}

%% file: tables/cifar100-superclass.tex
\begin{table}[h!]
\small
\centering
\setlength{\tabcolsep}{3pt}
\caption{\textbf{Superclasses for CIFAR100.} 
We document the members of all the superclasses provided by~\citet{krizhevsky2009learning}.
}
\label{table:cifar100-superclass}
\resizebox{\columnwidth}{!}{
\begin{tabular}{
    cll
}
\specialrule{.15em}{.05em}{.05em}
{$\gR$} & {\tt name} & $\gY_{\tt name}$
\\
\cmidrule{1-3}
0 &
aquatic mammals & beaver, dolphin, otter, seal, whale
\\
1 &
fish & aquarium fish, flatfish, ray, shark, trout
\\
2 &
flowers & orchid, poppy, rose, sunflower, tulip
\\
3 &
food containers & bottle, bowl, can, cup, plate
\\
4 &
fruit and vegetables & apple, mushroom, orange, pear, sweet pepper
\\
5 &
household electrical devices & clock, keyboard, lamp, telephone, television
\\
6 &
household furniture & bed, chair, couch, table, wardrobe
\\
7 &
insects & bee, beetle, butterfly, caterpillar, cockroach
\\
8 &
large carnivores & bear, leopard, lion, tiger, wolf
\\
9 &
large man-made outdoor things & bridge, castle, house, road, skyscraper
\\
10 &
large natural outdoor scenes & cloud, forest, mountain, plain, sea
\\
11 &
large omnivores and herbivores & camel, cattle, chimpanzee, elephant, kangaroo
\\
12 &
medium-sized mammals & fox, porcupine, possum, raccoon, skunk
\\
13 &
non-insect invertebrates & crab, lobster, snail, spider, worm
\\
14 &
people & baby, boy, girl, man, woman
\\
15 &
reptiles & crocodile, dinosaur, lizard, snake, turtle
\\
16 &
small mammals & hamster, mouse, rabbit, shrew, squirrel
\\
17 &
trees & maple tree, oak tree, palm tree, pine tree, willow tree
\\
18 &
vehicles 1 & bicycle, bus, motorcycle, pickup truck, train
\\
19 &
vehicles 2 & lawn mower, rocket, streetcar, tank, tractor
\\
\specialrule{.15em}{.05em}{.05em}
\end{tabular}
}
\end{table}

%% file: tables/sun397-superclass.tex
\begin{table}[h!]
\small
\centering
\setlength{\tabcolsep}{4pt}
\caption{\textbf{Superclasses for SUN397.} We document the members of the selected superclasses we used for SUN397~\cite{Xiao:2010} experiments. The classes that are not listed in this table are considered to be $\in \gR'$ in all SUN397 experiments.
}
\label{table:sun397-superclass}
\resizebox{\columnwidth}{!}{
\begin{tabular}{
    c
    l
    l
}
\specialrule{.15em}{.05em}{.05em}
{$\gR$} & {\tt name} & $\gY_{\tt name}$
\\
\cmidrule{1-3}
0 & 
\begin{tabular}[t]{l}
commercial buildings, \\ shops, markets, cities, \\ and towns 
\end{tabular}
& 
\begin{tabular}[t]{l}
alley, apartment building outdoor, building facade, diner outdoor, fire escape, fire station, \\ general store outdoor, hospital, hotel outdoor, inn outdoor, market outdoor, motel, office building, \\ phone booth, restaurant patio, schoolhouse, shopfront, skyscraper, slum, street, village
\end{tabular}
\\
\cmidrule{1-3}
1 & 
\begin{tabular}[t]{l}
cultural (art, education, \\religion, etc.)
\end{tabular}
& 
\begin{tabular}[t]{l}
apse indoor, aquarium, archive, art gallery, art school, art studio, auditorium, burial chamber, \\ catacomb, cathedral indoor, church indoor, classroom, cloister indoor, conference center, \\ courtroom, jail cell, jail indoor, kindergarden classroom, lecture room, library indoor, mosque indoor, \\ movie theater indoor, museum indoor, music studio, podium indoor, pulpit, stage indoor, \\ synagogue indoor, television studio, theater indoor procenium, theater indoor seats, throne room
\end{tabular}
\\
\cmidrule{1-3}
2 & 
\begin{tabular}[t]{l}
cultural or historical \\ building(place)
\end{tabular}
& 
\begin{tabular}[t]{l}
abbey, amphitheater, aqueduct, arch, basilica, bazaar outdoor, campus, castle, cathedral outdoor, \\ cemetery, church outdoor, courthouse, courtyard, fountain, kasbah, labyrinth outdoor, \\ library outdoor, lighthouse, mausoleum, medina, moat water, monastery outdoor, mosque outdoor, \\ oast house, observatory outdoor, pagoda, palace, planetarium outdoor, plaza, podium outdoor, ruin, \\ synagogue outdoor, temple east asia, temple south asia, tower, viaduct, windmill
\end{tabular}
\\
\cmidrule{1-3}
3 & 
\begin{tabular}[t]{l}
forest, field, jungle
\end{tabular}
& 
\begin{tabular}[t]{l}
bamboo forest, barn, botanical garden, corn field, corral, cottage garden, fairway, \\ field cultivated, field wild, forest broadleaf, forest needleleaf, forest path, forest road, \\ formal garden, golf course, hayfield, herb garden, orchard, outhouse outdoor, park, pasture, \\ picnic area, playground, putting green, rainforest, rice paddy, topiary garden, tree farm, \\ tree house, vegetable garden, vineyard, wheat field, wind farm, yard
\end{tabular}
\\
\cmidrule{1-3}
4 & 
\begin{tabular}[t]{l}
home or hotel
\end{tabular}
& 
\begin{tabular}[t]{l}
attic, basement, bathroom, bedroom, bow window indoor, childs room, closet, dinette home, \\ dining room, dorm room, game room, garage indoor, home office, hotel room, kitchen, kitchenette, \\ living room, nursery, pantry, parlor, playroom, poolroom home, recreation room, shower, \\ utility room, wet bar, youth hostel
\end{tabular}
\\
\cmidrule{1-3}
5 & 
\begin{tabular}[t]{l}
houses, cabins, \\
gardens, and farms
\end{tabular}
& 
\begin{tabular}[t]{l}
balcony exterior, balcony interior, barndoor, bow window outdoor, cabin outdoor, chalet, \\ chicken coop outdoor, doorway outdoor, driveway, greenhouse outdoor, house, \\ hunting lodge outdoor, kennel outdoor, mansion, manufactured home, patio, \\ residential neighborhood, shed, ski lodge, ski resort, veranda
\end{tabular}
\\
\cmidrule{1-3}
6 & 
\begin{tabular}[t]{l}
industrial and \\
construction
\end{tabular}
& 
\begin{tabular}[t]{l}
construction site, electrical substation, excavation, garbage dump, industrial area, landfill, \\ nuclear power plant outdoor, oil refinery outdoor, oilrig, power plant outdoor, trench,  water tower
\end{tabular}
\\
\cmidrule{1-3}
7 & 
\begin{tabular}[t]{l}
mountains, hills, \\
desert, sky
\end{tabular}
& 
\begin{tabular}[t]{l}
badlands, butte, canyon, cavern indoor, cliff, desert sand, desert vegetation, hill, mountain, \\ rock arch, sky, valley, volcano
\end{tabular}
\\
\cmidrule{1-3}
8 & 
\begin{tabular}[t]{l}
shopping and dining
\end{tabular}
& 
\begin{tabular}[t]{l}
bakery shop, banquet hall, bar, bazaar indoor, beauty salon, bistro indoor, bookstore, butchers shop, \\ cafeteria, candy store, clothing store, coffee shop, delicatessen, diner indoor,  discotheque, \\ drugstore, escalator indoor, fastfood restaurant, florist shop indoor, food court, general store indoor, \\ gift shop, ice cream parlor, jewelry shop, market indoor, music store, pharmacy, \\pub indoor, restaurant, restaurant kitchen, shoe shop, shopping mall indoor, supermarket, sushi bar, \\thriftshop, ticket booth, toyshop, videostore, wine cellar barrel storage, wine cellar bottle storage
\end{tabular}
\\
\cmidrule{1-3}
9 & 
\begin{tabular}[t]{l}
sports and leisure
\end{tabular}
& 
\begin{tabular}[t]{l}
amusement arcade, badminton court indoor, ball pit, ballroom, bowling alley, boxing ring, \\ casino indoor, firing range indoor, gymnasium indoor, ice skating rink indoor, jacuzzi indoor, \\ locker room, martial arts gym, poolroom establishment, riding arena, sauna, squash court, \\ swimming pool indoor, tennis court indoor, volleyball court indoor, wrestling ring indoor
\end{tabular}
\\
\specialrule{.15em}{.05em}{.05em}
\end{tabular}
}
\end{table}